\pdfoutput=1

\documentclass[11pt]{article}

\usepackage[final]{acl}

\usepackage{times}
\usepackage{latexsym}

\usepackage{amssymb}

\usepackage[T1]{fontenc}

\usepackage[utf8]{inputenc}

\usepackage{microtype}

\usepackage{array}

\usepackage{inconsolata}

\usepackage{graphicx}

\usepackage{amsfonts}
\usepackage{amsmath}

\title{CEMTM: Contextual Embedding-based Multimodal Topic Modeling}

\author{
    Amirhossein Abaskohi$^1$, Raymond Li$^1$, Chuyuan Li$^1$, Shafiq Joty$^2$, Giuseppe Carenini$^1$ \\ \\
    $^1$ Department of Computer Science The University of British Columbia \\
    V6T 1Z4, Vancouver, BC, Canada \\
    $^2$ Salesforce AI Research \\
    \texttt{\{aabaskoh, raymondl, carenini\}@cs.ubc.ca, chuyuan.li@ubc.ca} \\
    \texttt{sjoty@salesforce.com}
}

\begin{document}
\maketitle

\begin{abstract}
We introduce \textbf{CEMTM}, a context-enhanced multimodal topic model designed to infer coherent and interpretable topic structures from both short and long documents containing text and images. CEMTM builds on fine-tuned large vision language models (LVLMs) to obtain contextualized embeddings, and employs a distributional attention mechanism to weight token-level contributions to topic inference. A reconstruction objective aligns topic-based representations with the document embedding, encouraging semantic consistency across modalities. Unlike existing approaches, CEMTM can process multiple images per document without repeated encoding and maintains interpretability through explicit word-topic and document-topic distributions. Extensive experiments on six multimodal benchmarks show that CEMTM consistently outperforms unimodal and multimodal baselines, achieving a remarkable average LLM score of 2.61 (1-3 scale). Further analysis shows its effectiveness in downstream few-shot retrieval and its ability to capture visually grounded semantics in complex domains such as scientific articles\footnote{Code is publicly available at: \url{https://github.com/AmirAbaskohi/CEMTM}.}. 
\end{abstract}

\section{Introduction}

Topic modeling aims to uncover the latent thematic structure of a corpus by organizing documents into interpretable clusters of topics. While classical topic models like latent dirichlet allocation (LDA) \cite{blei2003latent} have long been applied to textual corpora, the rapid growth of multimodal content, where images, captions, and structured text co-exist, demands models that can jointly understand and reason over multiple modalities. Traditional multimodal topic models \cite{feng-lapata-2010-visual, 5540000} extended LDA to incorporate image features alongside text, but often failed to capture deeper cross-modal interactions. Recent advances in neural topic modeling \cite{10473197, gonzalez-pizarro-carenini-2024-neural} have addressed some of these limitations by learning shared embeddings across modalities, enabling more coherent and semantically unified topic discovery.

Parallel to these developments, large language models (LLMs) and large vision language models (LVLMs) have shown remarkable capacity to encode rich semantic knowledge from vast and diverse corpora. In text-based topic modeling, LLMs have been used both for generating and assigning topic with zero- and few-shot prompting \cite{mu-etal-2024-large, pham-etal-2024-topicgpt}, significantly improving topic coherence and interpretability. In multimodal settings, early efforts have used prompt-based methods \cite{prakash2023promptmtopic}. However, while models like TopicGPT produce interpretable outputs through natural language, they lack corpus-level topic distributions and robustness to prompt variation. They also do not model uncertainty or provide consistent global topic structures, limiting their usefulness for exploratory analysis \cite{10.1145/3623269}. A promising direction is to combine the knowledge grounding and modality alignment of LVLMs with the structured modeling of multimodal neural topic models, leveraging LVLMs to enhance semantic understanding without compromising the coherence and stability of topic representations. 

To address these limitations, we propose \textbf{CEMTM} (Contextual Embedding-based Multimodal Topic Modeling), a novel topic modeling framework that directly leverages the latent representations produced by pretrained LVLMs. Instead of designing complex architectures to align modalities, CEMTM uses the final token embedding from an LVLM as a compact, unified representation of a multimodal document that contains both textual and visual content \cite{jiang2025vlmvec}. This approach not only captures deeply aligned cross-modal semantics but also simplifies the processing of documents with multiple images. By avoiding the need for separate modality-specific encoders, CEMTM allows the entire document, including all images and the accompanying text, to be encoded holistically, making it well-suited for scalable and coherent multimodal topic modeling.  Additionally, inspired by \citet{fang-etal-2024-cwtm}, we incorporate a learnable importance network to estimate the contribution of each textual token and image patch to the document-topic representation. CEMTM achieves strong empirical performance across six benchmark datasets, obtaining an average LLM coherence score of \textbf{2.61} out of 3, outperforming a broad range of baselines.

Our contributions are: \textbf{(I)} We introduce \textbf{CEMTM}, a multimodal topic model that uses pretrained vision-language representations to generate coherent, diverse topics from long multimodal documents; \textbf{(II)} We propose a stochastic, distribution-based mechanism to learn token importance, improving semantic alignment and interpretability when combined with fine-tuned LVLM embeddings; \textbf{(III)} CEMTM sets a new SOTA, outperforming strong baselines on topic quality and the downstream task of few-shot question-answering (QA), demonstrating the value of topic distributions for retrieval-based tasks.

\section{Related Work}

\paragraph{Neural Multimodal Topic Modeling}
Early multimodal topic models extended LDA to handle image and text jointly \cite{blei2003modeling}, but often treated modalities independently. Neural approaches addressed this by learning shared representations, such as SupDocNADE \cite{zheng2014topic} and graph-based models for short documents \cite{10473197}. \citet{gonzalez2024neural} conducted a large-scale comparison of neural multimodal topic models, showing room for improvement in coherence and diversity. Unlike these models, CEMTM leverages pretrained LVLMs and uses their final token embeddings to capture aligned cross-modal semantics, eliminating the need to learn modality alignment during topic representation learning.

\paragraph{Language Models for Topic Modeling}
Language models have advanced topic modeling through prompting and contextual embeddings. Prompt-based methods like TopicGPT \citep{pham2024topicgpt} generate interpretable, natural-language topics with LLMs, while CWTM \citep{fang-etal-2024-cwtm} integrates contextual BERT embeddings into neural topic models for improved coherence. In multimodal settings, PromptMTopic \citep{prakash2023promptmtopic} combines textual and visual cues via LLMs to extract culturally aware topics from memes via prompting. More broadly, LVLMs offer unified representations for image–text pairs. However, their application to multimodal topic modeling remains underexplored. To address this, CEMTM leverages LVLMs, using the final token as a compact and aligned multimodal embedding, enabling efficient and interpretable topic discovery by using LLM's pretrained knowledge, without separate modality encoders or prompting.

\section{Method}

\begin{figure*}
    \centering
    \includegraphics[width=0.78\textwidth]{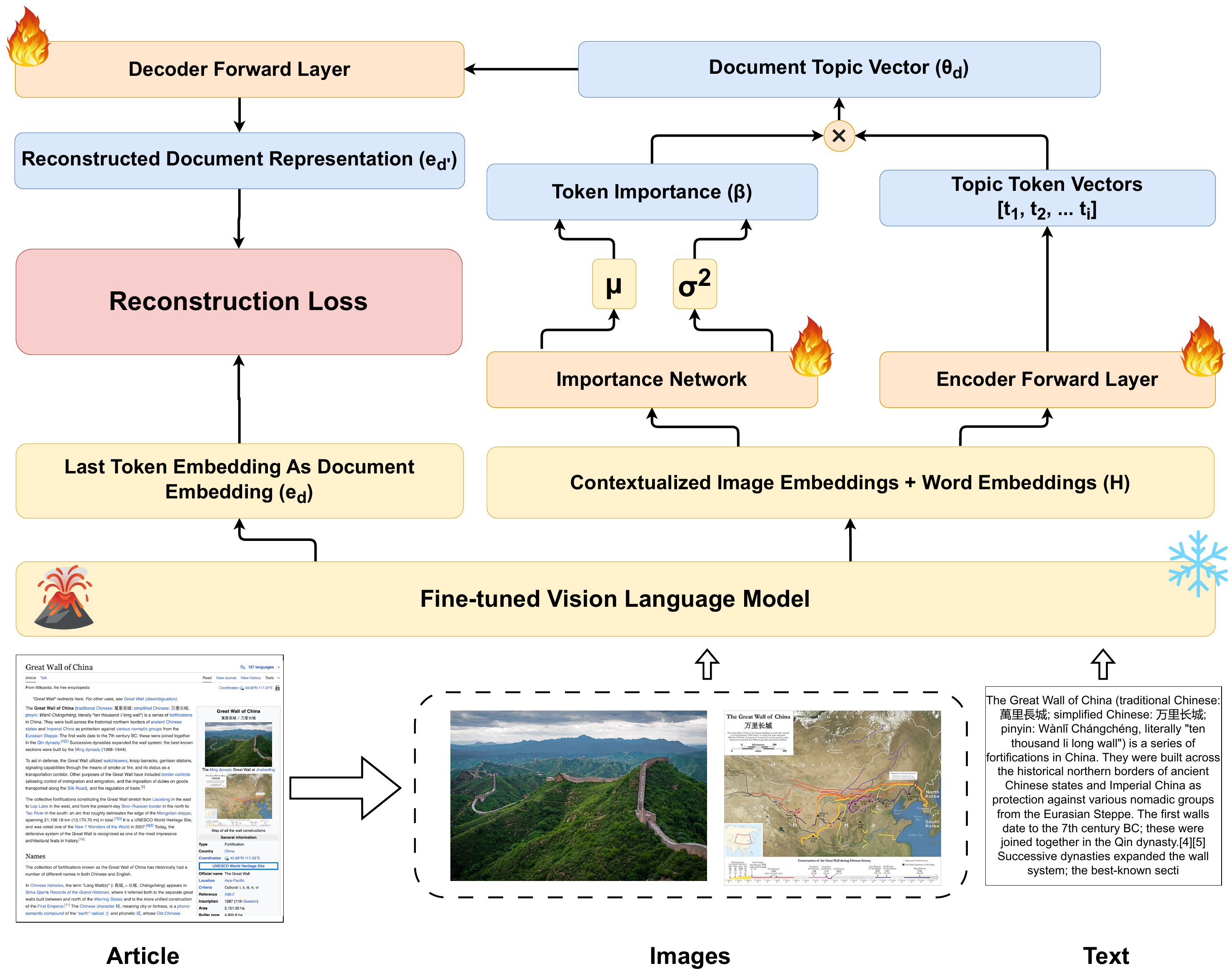}
    \caption{Overall architecture of CEMTM. Articles containing both text and images are encoded using a fine-tuned vision–language model to produce contextualized embeddings. During training, only the \textbf{decoder forward layer}, \textbf{encoder forward layer}, and \textbf{importance network} are fine-tuned, while the underlying vision–language backbone remains frozen. The model learns to construct document topic vectors by weighting token embeddings through the importance network, with reconstruction loss guiding optimization.}
    \label{fig:pipeline}
\end{figure*}

CEMTM is designed to perform soft topic modeling over long, multimodal documents. As shown in Figure \ref{fig:pipeline}, CEMTM processes both text and image inputs through an LVLM to produce contextualized token embeddings for both image and text tokens, learns importance-aware topic vectors, and reconstructs semantic document-level representations as supervision. We present our approach in three parts: document preprocessing (Section \ref{subsec:method_preprocess}), model training (Section \ref{subsec:method_training}), and topic extraction (Section \ref{subsec:method_extraction}).

\subsection{Preprocessing}
\label{subsec:method_preprocess}

Each document in the corpus contains both textual content and one or more associated images. Prior to training, we apply the following preprocessing steps. We begin with text cleaning, where we apply standard NLP preprocessing to remove punctuation, normalize casing, and eliminate irrelevant tokens (e.g., HTML tags). Following this, we perform vocabulary construction by tokenizing all documents and building a fixed vocabulary $\mathcal{V}$ that retains the most frequent words while discarding stop-words and rare terms. For the image processing step, all associated images are resized and formatted to ensure compatibility with the input requirements of the vision-language model.

\subsection{Model Training}
\label{subsec:method_training}

We use VLM2Vec \cite{jiang2025vlmvec}, a fine-tuned version of LLaVA-Next-7B \cite{liu2024llavanext}, to encode each document's text and image content into contextualized representations. Our approach is motivated by the hypothesis that while document embeddings encode rich semantic information, using them alone to infer topic distributions prevents access to vocabulary-level topic-word associations, limiting interpretability.

We begin by considering the approach of inferring latent document-topic vectors from document embeddings. Let $\mathbf{e}_d \in \mathbb{R}^D$ be the embedding of a document $d$ obtained from an LVLM, where $D$ denotes the dimensionality of the embedding space. A straightforward method would use the document embedding vector to generate the topics. However, this formulation lacks a way to associate topics with specific words, since it bypasses vocabulary-level granularity. To address this, we instead extract contextualized token embeddings from the document: 
\[
\mathbf{H} = [\mathbf{h}_1, \ldots, \mathbf{h}_N] \in \mathbb{R}^{N \times D}
\]
where $N$ is the number of textual tokens and visual patches in the document. Each $\mathbf{h}_i$ corresponds to a context-dependent representation of a token or an image patch. Each contextual embedding $\mathbf{h}_i$ is projected into the topic space using a \textbf{learnable encoder} with weight $\mathbf{W}_t \in \mathbb{R}^{D \times K}$, where $K$ denotes the number of latent topics, as follows:
\[
\mathbf{t}_i = Softmax(\mathbf{h}_i \mathbf{W}_t) \in \mathbb{R}^K
\]
We interpret $\mathbf{t}_i = p(z \mid \mathbf{h}_i)$ as the soft topic distribution for token $i$. However, not all tokens contribute equally to the semantic representation of a document. To model the relative importance of each token in shaping the document’s semantics, we introduce a \textbf{learnable} importance network that predicts a stochastic weight for each token. The importance network consists of a transformer encoder followed by a feedforward projection layer. Given contextualized token embeddings $\mathbf{H} = [\mathbf{h}_1, \ldots, \mathbf{h}_N]$, the importance network outputs a mean and standard deviation for each token’s importance score:
\[
\mu_i, \sigma^2_i = f_\theta(\mathrm{Transformer}(\mathbf{H}))_i
\]
\[
\alpha_i \sim \mathcal{N}(\mu_i, \sigma_i^2)
\]

To produce normalized importance weights, we apply a softmax across the sampled values:
\[
\boldsymbol{\beta} = Softmax([\alpha_1, \ldots, \alpha_N]) \in \mathbb{R}^N
\]

Advantageously,  the importance network also improves \textbf{interpretability}. The stochastic weights $\boldsymbol{\beta}$ provide an explicit estimate of how much each token or patch contributes to the document-level semantics. This makes it possible to identify salient tokens or regions and trace the evidence behind topic assignments, offering a more transparent view compared to standard attention mechanisms.

The document-topic vector is then computed by taking a weighted average of the token-level topic vectors:
\[
\boldsymbol{\theta}_d = Softmax(\sum_{i=1}^N \beta_i \mathbf{t}_i)
\]

The document-topic vector is then mapped into the embedding space using a \textbf{learnable decoder} with weight 
$\mathbf{W}_d \in \mathbb{R}^{K \times D}$, where $K$ is the number of latent topics and $D$ is the embedding dimension:
\[
e_{d'} = \boldsymbol{\theta}_d \mathbf{W}_d \in \mathbb{R}^D
\]

For supervision, we use the final token’s hidden state from VLM2Vec as the reference document embedding ($e_d$). The model maps the predicted embedding ($e_{d'}$) to this target, optimized with a reconstruction loss:
\[
\mathcal{L}_{\text{rec}} = \mathrm{MSE}(e_{d'}, e_d)
\]

This objective helps ensure that the learned topics preserve the global semantics encoded by the vision-language model, resulting in more coherent and multimodally grounded topic representations.

To encourage sharp and interpretable importance scores, we add an entropy regularization term to the loss \cite{vulic-mrksic-2018-specialising}. This term penalizes high-entropy (i.e., overly uniform) distributions over the importance weights \(\beta_i\), pushing the model to concentrate attention on a smaller subset of relevant elements. This promotes sparsity in the importance scores, making the model's decisions more focused and interpretable, which benefits both transparency and performance in reasoning tasks. The entropy regularization is defined as:
\[
\mathcal{L}_{\text{ent}} = \sum_{i=1}^{N} \beta_i \log \beta_i
\]

We also apply a KL divergence penalty between the predicted importance distribution $q(\alpha_i) = \mathcal{N}(\mu_i, \sigma_i^2)$ and a standard normal prior $p(\alpha_i) = \mathcal{N}(0, 1)$. This regularization keeps topic importance variables close to a standard Gaussian, reducing overfitting and promoting a smooth, balanced latent space \cite{jin-etal-2021-neural}. This is crucial in multimodal settings to avoid overconfident or modality-biased topic representations.
\[
\mathcal{L}_{\text{KL}} = \sum_{i=1}^N \left( \log \frac{1}{\sigma_i} + \frac{\sigma_i^2 + \mu_i^2 - 1}{2} \right)
\]

The final loss function is:
\[
\mathcal{L} = \mathcal{L}_{\text{rec}} + \lambda_{\text{ent}} \mathcal{L}_{\text{ent}} + \lambda_{\text{KL}} \mathcal{L}_{\text{KL}}
\]
where $\lambda_{\text{ent}}$ and $\lambda_{\text{KL}}$ are hyperparameters that control the strength of entropy and KL regularization, respectively.

This formulation enables the model to learn a flexible, distribution-based importance mechanism over tokens, while ensuring that the topic vector faithfully reconstructs document-level semantics and supports interpretable word-topic associations.

\subsection{Topic Extraction}
\label{subsec:method_extraction}

Once the model is trained, we extract topic-word associations by aggregating token-level topic vectors for each word in the vocabulary. Let $w \in \mathcal{V}$ be a word and $\mathcal{I}_w$ the set of all positions where $w$ appears in the corpus. We compute the aggregated topic vector for word $w$ as:
\[
\mathbf{t}_w = \frac{1}{Z_w} \sum_{i \in \mathcal{I}_w} \beta_i \mathbf{t}_i,
\]
where $Z_w = \sum_{i \in \mathcal{I}_w} \beta_i$ ensures normalization. The topic score for word $w$ in topic $k$ is $\mathbf{t}_w^{(k)}$, which is guaranteed to be non-negative due to the softmax used in the importance distribution. To extract representative topic words, we rank all words $w \in \mathcal{V}$ by their value $\mathbf{t}_w^{(k)}$ for each topic $k$. 

Topic words are extracted based on the vocabulary. For image patches, we associate each patch with its nearest word token in the embedding space and use that token in the aggregation step. This allows us to incorporate the semantic contribution of visual information while keeping the topic-word distributions interpretable.

\section{Experiments and Results}
We conduct extensive experiments to evaluate the effectiveness of our proposed model, CEMTM, on both topic modeling and its application to topic-guided few-shot retrieval for multimodal question answering. We assess the quality of the extracted topics using standard coherence and diversity metrics, and demonstrate the utility of the learned document-topic vectors in improving few-shot example selection. Additionally, we analyze the sensitivity of the model to the underlying encoder and provide qualitative insights into the learned topics and retrieval behavior. Refer to Appendix~\ref{sec:hyperparamter_settings} for hyperparameter and experimental settings.

\subsection{Datasets}

We evaluate CEMTM across a diverse set of multimodal and long-document datasets spanning encyclopedic, scientific, narrative, educational, and social domains. Table \ref{tab:dataset_stats} summarizes the datasets used in this study. Among these, only WikiWeb2M and SPIQA provide explicit ground-truth topic labels, which we use for quantitative evaluation. For the remaining datasets, we assess topic quality using unsupervised metrics such as coherence and diversity.

\begin{table}[!h]
    \centering
    \scriptsize
    \renewcommand{\arraystretch}{1.2}
    \resizebox{\linewidth}{!}{%
    \begin{tabular}{l|cccc}
        \hline
        \textbf{Dataset} & \textbf{Domain} & \textbf{\# Docs} & \textbf{Avg. Tokens} & \textbf{Avg. Images} \\
        \hline
        \textbf{WikiWeb2M} & Encyclopedic & 100,833 & 527 & 4.1 \\
        \textbf{SPIQA} & Scientific & 697 & 1342 & 3.7 \\
        \textbf{VIST} & Narrative & 50,000 & 152 & 5.0 \\
        \textbf{TQA} & Educational & 410 & 1086 & 2.9 \\
        \textbf{MSCOCO} & Image Captions & 30,000 & 13 & 1.0 \\
        \textbf{T4SA} & Social Media & 30,000 & 15 & 1.0 \\
        \textbf{FHM} & Memes & 10,000 & 9 & 1.0 \\
        \hline
    \end{tabular}
    }
    \caption{Summary of datasets used in our experiments.}
    \label{tab:dataset_stats}
\end{table}

\subsection{Evaluation Metrics}
We evaluate topic quality using five standard metrics: Normalized Pointwise Mutual Information (NPMI) \cite{lau2014machine}, Word Embedding score (WE) \cite{fang2016using}, LLM score \cite{stammbach2023revisiting}, Inverse Rank-Biased Overlap (I-RBO) \cite{terragni2021word}, and Topic Diversity (TD) \cite{dieng-etal-2020-topic}. NPMI and WE measure word-level coherence (co-occurrence and semantic similarity), the LLM score uses a language model to rate coherence (1–3 scale), and has strong correlation with human judgments \cite{stammbach2023revisiting}. I-RBO and TD capture diversity, respectively via rank-aware dissimilarity and unique word coverage.  

For datasets with gold topic labels (WikiWeb2M, SPIQA), we also report clustering-based metrics: Purity \cite{zhao2001criterion}, Adjusted Rand Index (ARI) \cite{hubert1985comparing}, and Normalized Mutual Information (NMI) \cite{strehl2002cluster}, which measure alignment between predicted and true topic assignments.  



\subsection{Baselines}

\begin{figure}[!t]
    \centering
    \includegraphics[width=0.6\linewidth]{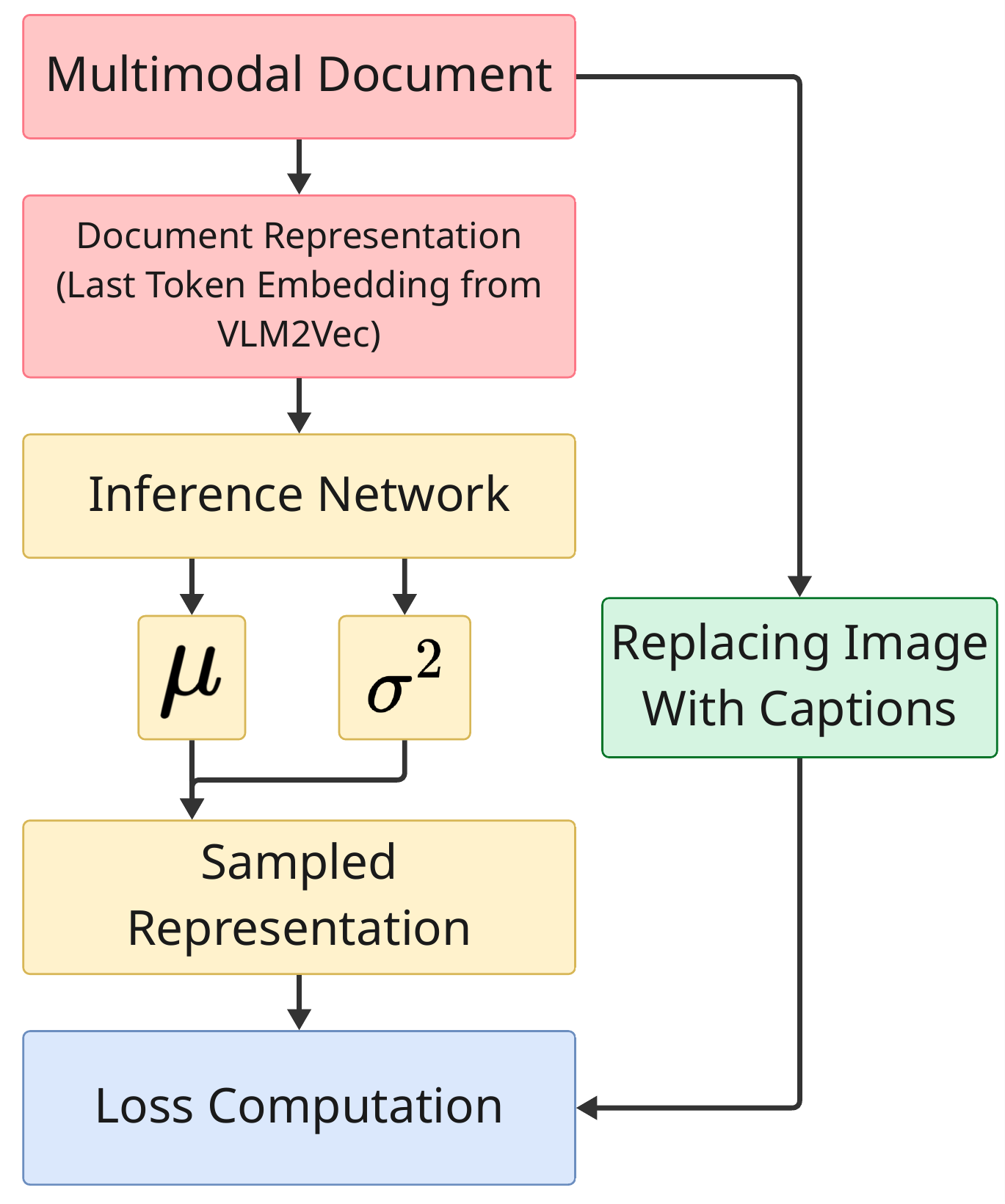}
    \caption{LVLM Zero-shot TM uses LVLM embeddings for better multimodal alignment and more meaningful topic vectors than Multimodal Zero-shot TM.}
    \label{fig:lvlm_zero_shot_tm}
\end{figure}

\begin{table*}[!t]
    \centering
    \renewcommand{\arraystretch}{1.2}
    \resizebox{\textwidth}{!}{%
        \begin{tabular}{l|ccccc|ccc|ccccc|ccc}
            \hline
                                             & \multicolumn{8}{c|}{\textbf{WikiWeb2M}}                                                                                                      & \multicolumn{8}{c}{\textbf{SPIQA}}                                                                                                           \\ \cline{2-17} 
                                             & NPMI            & WE              & LLM            & TD              & I-RBO           & Purity          & ARI             & NMI             & NPMI            & WE              & LLM            & TD              & I-RBO           & Purity          & ARI             & NMI             \\ \hline
            \textbf{LDA}                     & .028          & .095          & 2.40          & .703          & .953          & .295          & .131          & .235          & .022          & .088          & 2.31          & .717          & .942          & .299          & .136          & .244          \\
            \textbf{CombinedTM}              & .039          & .150          & 2.46          & .696          & .948          & .317          & .149          & .258          & .033          & .140          & 2.39          & .705          & .940          & .315          & .148          & .258          \\
            \textbf{Zero-shot TM}            & .040          & .172          & 2.51          & .717          & .966          & .335          & .149          & .257          & .036          & .162          & 2.46          & .731          & .958          & .331          & .152          & .263          \\
            \textbf{CWTM}                    & .052          & .188          & 2.56          & .714          & .965          & .347          & .167          & .275          & .047          & .177          & 2.51          & .729          & .957          & .344          & .168          & .278          \\
            \textbf{TopicGPT}                & .063          & .212          & 2.59          & .729          & -               & .378          & .189          & .288          & .057          & .201          & 2.55          & .748          & -               & .377          & .192          & .294          \\ \hline
            \textbf{M3L-Contrast}            & .065          & .226          & 2.62          & .744          & .981          & .386          & .196          & .298          & .059          & .215          & 2.59          & .763          & .973          & .387          & .199          & .304          \\
            \textbf{Multimodal Zero-shot TM} & .071          & .236          & 2.64          & .756          & .984               & .395          & .204          & .308          & .062          & .223          & 2.60          & .776          & .977               & .399          & .206          & .315          \\
            \textbf{LVLM Zero-shot TM}       & .074          & .246          & 2.65          & .763          & .990          & .407          & .213          & .320          & .065          & .233          & 2.63          & .785          & .980          & .411          & .215          & .326          \\
            \textbf{Multimodal TopicGPT}     & .080          & .255          & 2.67          & .774          & -          & .414          & .224          & .328          & .071          & .242          & 2.65          & .798          & -          & .419          & .227          & .335          \\ \hline
            \textbf{CEMTM (ours)}            & \textbf{.088} & \textbf{.272} & \textbf{2.70} & \textbf{.792} & \textbf{.996} & \textbf{.435} & \textbf{.245} & \textbf{.351} & \textbf{.080} & \textbf{.258} & \textbf{2.68} & \textbf{.817} & \textbf{.987} & \textbf{.444} & \textbf{.251} & \textbf{.359} \\ \hline
        \end{tabular}
    }
    \caption{Comparison of topic modeling performance on WikiWeb2M \cite{burns2023wiki} and SPIQA \cite{pramanick2024spiqa}. We report coherence (NPMI, WE, LLM), diversity (TD), redundancy (I-RBO), and clustering metrics (Purity, ARI, NMI), averaged over \( K = \{25, 50, 75, 100\} \) with three random seeds. CEMTM consistently outperforms all baselines. See Table \ref{table:detaile1} in Appendix \ref{sec:detailed_res} for detailed results for each K.}
    \label{table:qunatitative1}
\end{table*}

\begin{table*}[!t]
    \centering
    \renewcommand{\arraystretch}{1}
    \resizebox{0.85\textwidth}{!}{%
        \begin{tabular}{c}
            \begin{tabular}{l|ccccc|ccccc}
                \hline
                & \multicolumn{5}{c|}{\textbf{VIST}} & \multicolumn{5}{c}{\textbf{TQA}} \\
                \cline{2-11}
                & NPMI & WE & LLM & TD & I-RBO & NPMI & WE & LLM & TD & I-RBO \\
                \hline
                \textbf{LDA}                     & .017 & .077 & 2.23 & .646 & .935 & .019 & .081 & 2.25 & .665 & .940 \\
                \textbf{CombinedTM}              & .024 & .119 & 2.31 & .637 & .933 & .028 & .129 & 2.32 & .652 & .937 \\
                \textbf{Zero-shot TM}            & .029 & .138 & 2.38 & .659 & .949 & .032 & .151 & 2.39 & .679 & .955 \\
                \textbf{CWTM}                    & .036 & .155 & 2.44 & .656 & .946 & .041 & .169 & 2.45 & .675 & .953 \\
                \textbf{TopicGPT}                & .043 & .179 & 2.47 & .671 & -    & .050 & .194 & 2.48 & .692 & -    \\ \hline
                \textbf{M3L-Contrast}            & .044 & .190 & 2.50 & .681 & .962 & .052 & .207 & 2.51 & .705 & .970 \\
                \textbf{Multimodal Zero-shot TM} & .048 & .197 & 2.52 & .687 & .971 & .056 & .215 & 2.53 & .716 & .976 \\
                \textbf{LVLM Zero-shot TM}       & .050 & .208 & 2.54 & .696 & .974 & .059 & .226 & 2.55 & .724 & .977 \\
                \textbf{Multimodal TopicGPT}     & .055 & .216 & 2.56 & .707 & -    & .064 & .234 & 2.57 & .736 & -    \\ \hline
                \textbf{CEMTM (ours)}            & \textbf{.062} & \textbf{.233} & \textbf{2.58} & \textbf{.723} & \textbf{.981} & \textbf{.071} & \textbf{.250} & \textbf{2.60} & \textbf{.752} & \textbf{.984} \\
                \hline
            \end{tabular} \\
            \begin{tabular}{l|ccccc|ccccc}
                \hline
                & \multicolumn{5}{c|}{\textbf{MSCOCO}} & \multicolumn{5}{c}{\textbf{T4SA}} \\
                \cline{2-11}
                & NPMI & WE & LLM & TD & I-RBO & NPMI & WE & LLM & TD & I-RBO \\
                \hline
                \textbf{LDA}                     & .016 & .073 & 2.21 & .618 & .985 & .012 & .064 & 2.18 & .597 & .985 \\
                \textbf{CombinedTM}              & .023 & .117 & 2.28 & .605 & .984 & .018 & .105 & 2.25 & .585 & .978 \\
                \textbf{Zero-shot TM}            & .027 & .135 & 2.34 & .629 & .987 & .023 & .123 & 2.33 & .610 & .987 \\
                \textbf{CWTM}                    & .034 & .153 & 2.40 & .626 & .987 & .029 & .142 & 2.38 & .607 & .988 \\
                \textbf{TopicGPT}                & .042 & .177 & 2.43 & .642 & -    & .035 & .164 & 2.42 & .623 & -    \\ \hline
                \textbf{M3L-Contrast}            & .044 & .189 & 2.46 & .654 & .990 & .037 & .175 & 2.45 & .636 & .991 \\
                \textbf{Multimodal Zero-shot TM} & .047 & .198 & 2.48 & .662 & .992 & .040 & .182 & 2.46 & .644 & .992 \\
                \textbf{LVLM Zero-shot TM}       & .050 & .210 & 2.50 & .670 & .993 & .043 & .194 & 2.48 & .652 & .993 \\
                \textbf{Multimodal TopicGPT}     & .055 & .218 & 2.52 & .682 & -    & .048 & .202 & 2.50 & .663 & -    \\ \hline
                \textbf{CEMTM (ours)}            & \textbf{.061} & \textbf{.233} & \textbf{2.54} & \textbf{.697} & \textbf{.995} & \textbf{.053} & \textbf{.218} & \textbf{2.52} & \textbf{.679} & \textbf{.995} \\
                \hline
            \end{tabular}
        \end{tabular}
    }
    \caption{Unsupervised topic quality on VIST \cite{huang2016visual}, TQA \cite{kembhavi2017you}, MSCOCO \cite{lin2014microsoft}, and T4SA \cite{Vadicamo_2017_ICCVW} using coherence (NPMI, WE, LLM), diversity (TD), and redundancy (I-RBO). Results are averaged over \( K = \{25, 50, 75, 100\} \) with three random seeds. CEMTM outperforms all baselines. See Table \ref{table:detaile2} in Appendix \ref{sec:detailed_res} for detailed results for each K. }
    \label{table:qunatitative2}
\end{table*}

We compare CEMTM against a broad set of baselines spanning traditional, contextualized, and multimodal topic modeling. As a classical reference, Latent Dirichlet Allocation (LDA) \cite{blei2003latent}, trained with Gensim \cite{rehurek2010software}, models each document as a mixture of topics over a bag-of-words representation. More recent contextualized methods replace or augment bag of words (BoW) features with embeddings: ZeroshotTM \cite{bianchi2021pre} uses SBERT embeddings \cite{reimers-gurevych-2019-sentence} for zero-shot topic modeling, and CombinedTM \cite{bianchi-etal-2021-cross} improves interpretability by concatenating SBERT with BoW. Similarly, CWTM \cite{fang-etal-2024-cwtm} projects contextual token representations into a topic space and aggregates them using fixed or learned importance scores.  

We also adapt TopicGPT \cite{pham2024topicgpt}, which does not expose explicit topic-word distributions. To approximate them, we restrict the number of topics to $K$ and use token-level soft assignments to construct interpretable topic-word vectors (see Appendix \ref{sec:hyperparamter_settings} for more details). Building on this, we introduce Multimodal TopicGPT, which incorporates both text and images at inference.  

For other multimodal baselines, M3L-Contrast \cite{zosa-pivovarova-2022-multilingual} leverages image–caption alignment to enforce consistent document-topic representations, while Multimodal ZeroshotTM \cite{gonzalez-pizarro-carenini-2024-neural} extends ZeroshotTM by combining textual embeddings with vision encoder features. Finally, our proposed LVLM ZeroshotTM strengthens this approach by using embeddings from LVLMs, yielding more semantically grounded and better aligned multimodal topic vectors (as shown in Figure \ref{fig:lvlm_zero_shot_tm}).

\subsection{Quantitative Results}

We evaluate the performance of CEMTM and baselines across a wide range of datasets, reporting both intrinsic topic quality metrics (e.g., NPMI, WE, LLM, TD, I-RBO) and extrinsic clustering metrics (Purity, ARI, NMI) when ground-truth labels are available. Results are averaged over four topic counts ($K = 25, 50, 75, 100$), each run with three random seeds.

\paragraph{Long-document and In Domain Performance.} Table~\ref{table:qunatitative1} presents results on WikiWeb2M and SPIQA, both of which consist of long, multimodal documents and include ground-truth topic annotations. CEMTM outperforms all baselines across every metric, demonstrating stronger topic coherence, higher diversity, and more accurate topic assignments. Notably, our model surpasses multimodal baselines like Multimodal TopicGPT and LVLM Zero-shot TM, while also being more efficient than methods like TopicGPT that require autoregressive decoding or multiple forward passes (for topic generation and topic assignment). Unlike other models, CEMTM processes documents with multiple images in a single pass without repeated inference, offering both performance and scalability benefits.

\paragraph{Generalization Across Domains.} Table~\ref{table:qunatitative2} shows performance on four additional datasets—VIST, TQA, MSCOCO, and T4SA—that include both short and medium-length multimodal documents but lack ground-truth topic labels. Again, CEMTM achieves the best performance across all intrinsic metrics and datasets, highlighting its flexibility across domains including narratives (VIST), educational content (TQA), captioned images (MSCOCO), and social media posts (T4SA). These results indicate that the model generalizes well even beyond long-text scenarios.

\begin{table}[!t]
    \renewcommand{\arraystretch}{1.1}
    \resizebox{\linewidth}{!}{%
        \begin{tabular}{l|ccccc}
            \hline
                                             & \multicolumn{5}{c}{\textbf{FHM}}                                                      \\ \cline{2-6} 
                                             & NPMI            & WE              & LLM             & TD              & I-RBO           \\ \hline
            \textbf{LDA}                     & .005          & .048          & 2.04          & .530          & .983          \\
            \textbf{CombinedTM}              & .009          & .087          & 2.10          & .518          & .975          \\
            \textbf{Zero-shot TM}            & .014          & .109          & 2.17          & .543          & .984          \\
            \textbf{CWTM}                    & .019          & .128          & 2.22          & .540          & .986          \\
            \textbf{TopicGPT}                & .025          & .150          & 2.26          & .554          & -          \\ \hline
            \textbf{M3L-Contrast}            & .030          & .169          & 2.34          & .566          & .990          \\
            \textbf{Multimodal Zero-shot TM} & .033          & .177          & 2.36          & .574          & .992          \\
            \textbf{LVLM Zero-shot TM}       & .039          & .194          & 2.43          & .590          & .993          \\
            \textbf{Multimodal TopicGPT}     & .043          & .202          & 2.45          & .601          & -          \\ \hline
            \textbf{CEMTM (ours)}            & \textbf{.049} & \textbf{.217} & \textbf{2.47} & \textbf{.617} & \textbf{.995} \\ \hline
        \end{tabular}
    }
    \caption{Unsupervised topic quality on the FHM dataset \cite{10.5555/3495724.3495944}, which tests modeling under high image–text semantic gaps. We report coherence (NPMI, WE, LLM), diversity (TD), and redundancy (I-RBO), averaged over \( K = \{25, 50, 75, 100\} \) with three seeds. CEMTM outperforms all baselines, highlighting the benefit of joint multimodal modeling. See Table \ref{table:detailed3} in Appendix \ref{sec:detailed_res} for detailed results for each K.  }
    \label{table:qunatitative3}
\end{table}

\paragraph{Semantic Gap Analysis.} A particularly challenging scenario in multimodal topic modeling arises when there is a semantic gap between images and text, as in the case of memes. Table~\ref{table:qunatitative3} focuses on the Facebook Hateful Memes dataset, where there is a known semantic gap between images and their accompanying captions. This setting is particularly challenging for topic models that rely on textual content alone. The results show a clear separation between unimodal and multimodal models, with image-aware approaches consistently outperforming text-only counterparts. Furthermore, models that use large vision-language models (LVLMs), such as LVLM Zero-shot TM, Multimodal TopicGPT, and CEMTM, show the highest gains, suggesting that better multimodal alignment significantly improves topic modeling in semantically ambiguous contexts. This validates the design of CEMTM, which leverages fine-tuned LVLM embeddings and a flexible importance-weighted fusion mechanism to capture cross-modal semantics effectively.


\begin{table}[!t]
    \centering
    \renewcommand{\arraystretch}{1.2}
    \resizebox{\linewidth}{!}{%
    \begin{tabular}{l|cc|cc}
        \hline
        \textbf{Setting} & \multicolumn{2}{c|}{\textbf{SPIQA}} & \multicolumn{2}{c}{\textbf{TQA}} \\
         & METEOR & BERTScore-F1 & Acc & F1-Macro \\
        \hline
        \textbf{Zero-shot} & 26.3 & 67.48 & 84.87 & 83.79 \\ \hline
        \textbf{3-shot Random Selection} & 27.4 & 68.92 & 85.36 & 84.28 \\
        \textbf{3-shot Embedding Based Selection} & 28.7 & 70.11 & 86.09 & 85.12 \\ \hline
        \textbf{3-shot Topic Based Selection} & \textbf{31.3} & \textbf{72.76} & \textbf{87.31} & \textbf{87.03} \\
        \hline
    \end{tabular}
    }
    \caption{Few-shot QA results on SPIQA and TQA test sets. Topic-based selection leads to the best performance across both datasets. For a detailed comparison of the performance of different topic models used for topic-based retrieval, refer to Table~\ref{tab:retrieval_detailed} in Appendix~\ref{sec:retrieval_compare_models}.}
    \label{table:qa}
\end{table}

\begin{table*}[h]
    \centering
    \small
    \renewcommand{\arraystretch}{1.2}
    \resizebox{\linewidth}{!}{%
    \begin{tabular}{l|p{6cm}|p{6cm}}
    \hline
    \textbf{Model} & \textbf{Topic Words Inferred from Text} & \textbf{Topic Words Inferred from Visual Patches} \\
    \hline
    CWTM & eruption, magma, lava, ash, rock & --- \\
    Multimodal Zero-shot TM & eruption, lava, ash, magma, crater & plume, smoke, debris, slope, flow \\
    LVLM Zero-shot TM & eruption, lava, ash, magma, vent & cloud, plume, lava flow, crater rim, tephra \\
    CEMTM (ours) & eruption, lava, ash, magma, pyroclastic & plume, flow, cloud, fountain, tephra \\
    \hline
    \end{tabular}
    }
    \caption{Predicted top topic words for the Wikipedia page \textit{Volcanic eruption}, separated into text-derived vs. visual patch-derived words. The text columns intentionally show high overlap (e.g., \textit{eruption}, \textit{lava}, \textit{ash}, \textit{magma}) to reflect consistent lexical signals across models, with CEMTM additionally capturing specialized geological terminology such as \textit{pyroclastic}. The visual columns highlight model-specific perceptual cues (e.g., \textit{plume}, \textit{flow}, \textit{cloud}), showing how multimodal integration introduces eruption-specific semantics not present in text alone.}
    \label{tab:eruption_topics}
\end{table*}

\begin{table*}[h]
    \centering
    \small
    \renewcommand{\arraystretch}{1.15}
    \begin{tabular}{p{4cm}|p{11cm}}
    \hline
    \textbf{Query Page:} \textit{Saturn (planet)} &
    \textbf{Top Topic Words:} planet, ring, gas, orbit, atmosphere, moon, giant, solar, space, rotation \\
    \hline
    \textbf{Random} & \textit{Barack Obama}, \textit{Photosynthesis}, \textit{Succulent plant} \\
    \textbf{Embedding-based} & \textit{Solar System}, \textit{Mars}, \textit{Astronomy} \\
    \textbf{Topic-based (CEMTM)} & \textit{Jupiter}, \textit{Uranus}, \textit{Gas giant} \\
    \hline
    \textbf{Query Page:} \textit{French Revolution} &
    \textbf{Top Topic Words:} revolution, france, king, monarchy, liberty, citizens, republic, uprising, power, 1789 \\
    \hline
    \textbf{Random} & \textit{Harry Potter}, \textit{Mount Everest}, \textit{DNA replication} \\
    \textbf{Embedding-based} & \textit{American Revolution}, \textit{Napoleon}, \textit{History of France} \\
    \textbf{Topic-based (CEMTM)} & \textit{Reign of Terror}, \textit{Louis XVI}, \textit{Constitution of 1791} \\
    \hline
    \textbf{Query Page:} \textit{Photosynthesis} &
    \textbf{Top Topic Words:} plant, sunlight, chlorophyll, carbon, dioxide, glucose, energy, leaf, oxygen, process \\
    \hline
    \textbf{Random} & \textit{World War II}, \textit{Twitter}, \textit{Rome} \\
    \textbf{Embedding-based} & \textit{Cellular respiration}, \textit{Chloroplast}, \textit{Botany} \\
    \textbf{Topic-based (CEMTM)} & \textit{Light-dependent reactions}, \textit{Carbon fixation}, \textit{Thylakoid} \\
    \hline
    \end{tabular}
    \caption{Comparison of retrieval methods for Wikipedia pages. CEMTM yields more fine-grained, thematically aligned results by leveraging interpretable topic distributions.}
    \label{tab:retrieval_examples}
\end{table*}

\subsection{Improving Few-Shot Multimodal QA with Topic-Aware Retrieval}

Beyond evaluating CEMTM on topic modeling tasks, we assess the utility of its learned document-topic vectors for improving few-shot multimodal question answering. Specifically, we use these topic vectors (with the number of topics set to \( K = 50 \)) to retrieve in-context examples for prompting a QA model in a few-shot setting. We compare four retrieval strategies on the SPIQA and TQA test sets: (1) a zero-shot baseline, (2) random selection of 3 in-context examples, (3) embedding-based retrieval using cosine similarity over OpenAI's \texttt{text-embedding-3-small}\footnote{\url{https://platform.openai.com/docs/models/text-embedding-3-small}}, and (4) our topic-based retrieval using document-topic vectors produced by CEMTM. As shown in Table~\ref{table:qa}, topic-based selection significantly outperforms all other methods across all evaluation metrics, including METEOR and BERTScore on SPIQA, and accuracy and macro-F1 on TQA. This demonstrates that topic distributions learned by CEMTM capture high-level semantic structure that can guide effective example selection, providing relevant and diverse context without relying on direct surface similarity. These results highlight the potential of CEMTM beyond topic interpretability.

\subsection{Qualitative Results}

To further evaluate how CEMTM captures visually grounded semantics, we examine the Wikipedia article titled \textit{Volcanic eruption}, which describes types of volcanic eruptions, geological processes, and associated hazards. The page includes key images such as eruption plumes, lava flows, and ash clouds that visually differentiate between explosive and effusive eruptions, information that is often only implicitly mentioned or not described in detail in the text. Table~\ref{tab:eruption_topics} presents a comparison of top topic words predicted by CWTM (text-only), Multimodal Zero-shot TM, LVLM Zero-shot TM, and CEMTM. The text-only model generates general geological terms and omits eruption-specific visual cues. Multimodal Zero-shot TM incorporates visual features but lacks deep integration, leading to less coherent topic-word clusters. LVLM Zero-shot TM improves topic specificity, capturing visual elements like “plume” and “cloud,” while CEMTM further refines this by predicting visually aligned and geologically grounded terms (e.g. “pyroclastic”). CEMTM benefits from fine-grained fusion of text and image semantics during training, and its reconstruction objective ensures visual information is preserved in the topic structure, something BoW-based models discard. 


Table~\ref{tab:retrieval_examples} qualitatively illustrates how CEMTM enhances semantic retrieval by leveraging interpretable document-topic vectors. For each query Wikipedia article, CEMTM retrieves thematically precise pages by comparing topic distributions, outperforming both random and embedding-based baselines. While embedding-based methods retrieve broadly related pages (e.g., \textit{Mars} for \textit{Saturn}), they often lack topical granularity. In contrast, CEMTM identifies highly specific, contextually aligned documents such as \textit{Gas giant} or \textit{Constitution of 1791}, grounded in the core semantic fields of the queries. This demonstrates that topic-based retrieval with CEMTM not only captures more interpretable signals but also better models thematic structure, making it particularly useful for few-shot prompting and corpus exploration.

\begin{table*}[!t]
    \renewcommand{\arraystretch}{1.2}
    \resizebox{\textwidth}{!}{%
        \begin{tabular}{l|ccccc|ccc|ccccc|ccc}
            \hline
                                                                 & \multicolumn{8}{c|}{\textbf{WikiWeb2M}}                                                                                       & \multicolumn{8}{c}{\textbf{SPIQA}}                                                                                            \\ \cline{2-17} 
                                                                 & NPMI          & WE            & LLM           & TD            & I-RBO         & Purity        & ARI           & NMI           & NPMI          & WE            & LLM           & TD            & I-RBO         & Purity        & ARI           & NMI           \\ \hline
            \textbf{CEMTM}                                       & \textbf{.088} & \textbf{.272} & \textbf{2.70} & \textbf{.792} & \textbf{.996} & \textbf{.435} & \textbf{.245} & \textbf{.351} & \textbf{.080} & \textbf{.258} & \textbf{2.68} & \textbf{.817} & \textbf{.987} & \textbf{.444} & \textbf{.251} & \textbf{.359} \\ \hline
            \textbf{Without Distribution As Importance Netowork} & \underline{.087} & \underline{.269} & \underline{2.69} & \underline{.789} & \underline{.996} & \underline{.432} & \underline{.242} & \underline{.348} & \underline{.078} & \underline{.255} & \underline{2.68} & \underline{.814} & \underline{.987} & \underline{.441} & \underline{.248} & \underline{.356} \\ \hline
            \textbf{No VLM2Vec}                                  & .083          & .260          & 2.67          & .776          & .994          & .424          & .231          & .335          & .074          & .246          & 2.66          & .797          & .985          & .429          & .235          & .342          \\
            \textbf{VLM2Vec only for Word Embedding}             & \underline{.085} & .265          & \underline{2.68} & .780          & .994          & .426          & .234          & .338          & .075          & .249          & \underline{2.67} & .801          & .986          & .432          & .239          & .346          \\
            \textbf{VLM2Vec only for Document Embedding}         & \underline{.085} & \underline{.266} & \underline{2.68} & \underline{.781} & \underline{.995} & \underline{.428} & \underline{.235} & \underline{.340} & \underline{.076} & \underline{.251} & \underline{2.67} & \underline{.802} & \underline{.986} & \underline{.434} & \underline{.240} & \underline{.348} \\ \hline
        \end{tabular}
    }
    \caption{Ablation results on WikiWeb2M and SPIQA, showing the impact of using distribution-based importance modeling and fine-tuned VLM2Vec embeddings for word and document representations.}
    \label{table:ablations}
\end{table*}

\begin{table*}[!t]
    \renewcommand{\arraystretch}{1.1}
    \resizebox{\textwidth}{!}{%
        \begin{tabular}{l|ccccc|ccc|ccccc|ccc}
            \hline
                                                                 & \multicolumn{8}{c|}{\textbf{WikiWeb2M}}                                                                                       & \multicolumn{8}{c}{\textbf{SPIQA}}                                                                                            \\ \cline{2-17} 
                                                                 & NPMI          & WE            & LLM           & TD            & I-RBO         & Purity        & ARI           & NMI           & NPMI          & WE            & LLM           & TD            & I-RBO         & Purity        & ARI           & NMI           \\ \hline
            \textbf{CEMTM}                                & \textbf{.088} & \textbf{.272} & \textbf{2.70} & \textbf{.792} & \textbf{.996} & \textbf{.435} & \textbf{.245} & \textbf{.351} & \textbf{.080} & \textbf{.258} & \textbf{2.68} & \textbf{.817} & \textbf{.987} & \textbf{.444} & \textbf{.251} & \textbf{.359} \\ \hline
            \textbf{CEMTM w/o Image}                             & .070          & .228          & 2.64          & .765          & .981          & .398          & .205          & .310          & .062          & .215          & 2.62          & .785          & .973          & .400          & .207          & .318          \\
            \textbf{CEMTM w/ Caption}                           & .077          & .246          & 2.66          & .778          & .988          & .417          & .221          & .332          & .068          & .235          & 2.64          & .800          & .980          & .425          & .229          & .340          \\ \hline
        \end{tabular}
    }
    \caption{Modality ablation results on WikiWeb2M and SPIQA. 
    \textbf{CEMTM w/o Image} indicates removing the image modality entirely, 
    while \textbf{CEMTM w/ Caption} replaces the image modality with GPT-4o generated captions.}
    \label{table:modality_ablations}
\end{table*}

\section{Ablation Studies}


\subsection{Impact of Vision-Language Embedding Quality}
To assess the effect of vision-language pretraining and fine-tuning, we compare several variants that adjust how VLM2Vec is used in CEMTM. As shown in Table~\ref{table:ablations},  replacing the VLM2Vec version of LLaVA-Next-7B model, obtained by fine-tuning LLaVA-Next-7B, entirely with pretrained LLaVA-Next-7B results in the largest performance drop, particularly in document clustering metrics. This confirms that alignment-aware fine-tuned embeddings are crucial for accurate topic representation. Using VLM2Vec only for token embeddings or only for document embeddings results in intermediate performance: both help individually, but full use of VLM2Vec (as in the original model) provides the strongest gains. These results highlight the importance of semantically aligned, multimodal representations at both word and document levels. We further investigated the sensitivity of CEMTM across different LVLMs 
in Appendix~\ref{sec:encoding_sensitivity}.

\subsection{Role of Distributional Supervision in the Importance Network}
We further evaluate the effect of modeling importance weights as samples from a learned Gaussian distribution, rather than as deterministic values. As shown in Table~\ref{table:ablations}, removing this distributional supervision and replacing it with a simple softmax network leads to a consistent drop in performance across coherence (NPMI, WE, LLM), diversity (TD), and clustering metrics (Purity, ARI, NMI). This confirms that stochastic importance modeling not only improves robustness, but also helps the model better focus on semantically relevant tokens or image regions, ultimately yielding higher-quality and more interpretable topic structures.

\subsection{Role of Visual Signals in Multimodal Topic Modeling}
Table~\ref{table:modality_ablations} presents the results of our modality ablation study on WikiWeb2M and SPIQA. We observe that removing the image modality (\textbf{w/o Image}) substantially degrades performance across all metrics, confirming the crucial role of visual signals in enhancing topic coherence and clustering quality. When substituting images with automatically generated captions (\textbf{w/ Caption}), the performance improves compared to removing images entirely, but it still falls short of the full model (\textbf{CEMTM}). This finding is consistent with prior work emphasizing that captions only provide partial information about visual content, whereas direct image features capture richer multimodal cues. Overall, these results highlight the importance of incorporating image representations directly, rather than relying solely on textual surrogates.

\section{Conclusion}
We presented \textbf{CEMTM}, an interpretable multimodal topic model designed to extract coherent topics from both short and long documents containing text and images. CEMTM leverages fine-tuned LVLM embeddings alongside a distributional attention mechanism, combining contextualized representations with a reconstruction-based training objective and importance-weighted fusion. This enables the model to capture document-level semantics while preserving interpretability. Evaluated on six benchmark datasets, CEMTM achieves a strong average LLM score of 2.61 out of 3 and a Purity score of 0.44, outperforming a broad range of unimodal and multimodal baselines. Ablation results further highlight the value of fine-tuned LVLMs and distributional supervision in guiding topic quality. Overall, CEMTM is a scalable, explainable framework that enables tasks like few-shot retrieval, multimodal summarization, and corpus-level topic analysis with efficiency and interpretability.

\section*{Limitations}
While CEMTM demonstrates strong performance and scalability across diverse multimodal datasets, several limitations remain. First, the model relies heavily on pretrained LVLMs, which introduces significant computational overhead and requires access to large-scale GPU resources (See Appendix \ref{sec:hyperparamter_settings} for more information). This may limit the applicability of CEMTM in low-resource or real-time settings. 
Second, although the reconstruction objective aligns topic vectors with semantic document embeddings, this does not guarantee that each topic is fully disentangled or interpretable in isolation--particularly when documents cover overlapping concepts or when visual information is noisy or redundant. 
Additionally, our evaluation focuses on English-language datasets and does not explore multilingual or cross-cultural settings, where visual semantics and topic interpretability may differ significantly. 
Lastly, while the importance network encourages interpretability through attention sparsity, its learned weights are not explicitly validated against human judgments, leaving room for future work in explainability and user-in-the-loop topic refinement.

\section*{Ethical Considerations}
\paragraph{Potential Risks} This research presents potential risks related to the use of real-world multimodal data, which may contain harmful biases or inaccuracies. To mitigate these risks, all experiments were conducted in controlled settings, and none of the resulting models were deployed in public-facing systems. Additionally, we carefully monitored model outputs during evaluation to ensure that no harmful content was propagated.

\paragraph{FHM Offensive Data} We used the Facebook Hateful Memes (FHM) dataset, which contains potentially offensive content, strictly for experimental purposes in this study. To minimize harm, we do not release any models trained on this dataset. This precaution ensures that any biased or harmful patterns present in the data are not disseminated or used beyond the limited scope of our research. 

\paragraph{AI Assistance} AI tools were used during this project to assist with both writing and coding. Specifically, AI assistance supported drafting text, refining code structure, and improving clarity. However, all scientific contributions, including experimental design, analysis, and interpretation, were solely conducted by the authors to preserve research integrity.

\section*{Acknowledgements}
The authors thank the Natural Sciences and Engineering Research Council of Canada (NSERC) for their support of this research.

Nous remercions le Conseil de recherches en sciences naturelles et en génie du Canada (CRSNG) de son soutien.

\clearpage

\bibliography{custom}

\clearpage

\appendix

\section{Encoding Model Sensitivity }
\label{sec:encoding_sensitivity}

\begin{table*}[!t]
    \renewcommand{\arraystretch}{1.2}
    \resizebox{\textwidth}{!}{%
        \begin{tabular}{l|c|ccccc|ccc|ccccc|ccc}
\hline
                       & \textbf{VLM2Vec}    & \multicolumn{8}{c|}{\textbf{WikiWeb2M}}                                                                                       & \multicolumn{8}{c}{\textbf{SPIQA}}                                                                                            \\ \cline{3-18} 
                       & \textbf{Fine Tuned} & NPMI          & WE            & LLM           & TD            & I-RBO         & Purity        & ARI           & NMI           & NPMI          & WE            & LLM           & TD            & I-RBO         & Purity        & ARI           & NMI           \\ \hline
\textbf{LLava-Next-7B} & $\checkmark$        & .087          & .269          & 2.69          & .789          & .996          & .432          & .242          & .348          & .080          & .258          & 2.68          & .817          & .987          & .444          & .251          & .359          \\
\textbf{QWen2VL-7B}    & $\checkmark$        & \textbf{.093} & \textbf{.280} & \textbf{2.72} & \textbf{.796} & \textbf{.997} & \textbf{.444} & \textbf{.254} & \textbf{.361} & \textbf{.084} & \textbf{.269} & \textbf{2.71} & \textbf{.824} & \textbf{.991} & \textbf{.459} & \textbf{.263} & \textbf{.371} \\ \hline
\textbf{Phi-3.5-V}     & $\times$            & .083          & .255          & 2.67          & .777          & .993          & .414          & .228          & .332          & .074          & .244          & 2.65          & .804          & .984          & .425          & .235          & .342          \\
\textbf{LLaVA-1.6-7B}  & $\times$            & {\underline{.088}}    & {\underline{.272}}    & {\underline{2.70}}    & {\underline{.792}}    & {\underline{.996}}    & {\underline{.435}}    & {\underline{.245}}    & {\underline{.351}}    & {\underline{.080}}    & {\underline{.258}}    & {\underline{2.68}}    & {\underline{.817}}    & {\underline{.987}}    & {\underline{.444}}    & {\underline{.251}}    & {\underline{.359}}    \\
\textbf{CLIP}          & $\times$            & .081          & .251          & 2.67          & .774          & .991          & .410          & .222          & .328          & .072          & .241          & 2.65          & .800          & .982          & .421          & .231          & .339          \\
\textbf{BLIP2-OPT-7B}  & $\times$            & .083          & .255          & 2.67          & .777          & .993          & .414          & .228          & .332          & .074          & .244          & 2.65          & .805          & .984          & .425          & .235          & .343          \\ \hline
\end{tabular}
    }
    \caption{Impact of the underlying LVLM encoder on CEMTM performance. We compare LoRA fine-tuned vision-language models—LLaVA-Next-7B, LLaVA-1.6-7B, QWen2VL-7B, Phi-3.5-V, CLIP, and BLIP—as backbone encoders for CEMTM. Results are reported on WikiWeb2M and SPIQA across topic coherence (NPMI, WE, LLM).}
    \label{table:encoding_model_sensitivity}
\end{table*}

\begin{table*}[!t]
\centering
\renewcommand{\arraystretch}{1.1}
\resizebox{0.9\textwidth}{!}{%
\begin{tabular}{l|cc|cc}
\hline
\textbf{Setting} & \multicolumn{2}{c|}{\textbf{SPIQA}} & \multicolumn{2}{c}{\textbf{TQA}} \\
 & METEOR & BERTScore-F1 & Accuracy & F1-Macro \\
\hline
Zero-shot & 26.3 & 67.48 & 84.87 & 83.79 \\ \hline
3-shot Random Selection & 27.4 & 68.92 & 85.36 & 84.28 \\ 
3-shot Embedding-Based Selection & 28.7 & 70.11 & 86.09 & 85.12 \\ \hline
3-shot Topic-Based (CWTM) & 28.3 & 69.85 & 85.82 & 84.96 \\
3-shot Topic-Based (M3L-Contrast) & 28.9 & 70.22 & 86.09 & 85.18 \\
3-shot Topic-Based (Multimodal Zero-shot TM) & 29.4 & 70.63 & 86.23 & 85.43 \\
3-shot Topic-Based (LVLM Zero-shot TM) & 29.8 & 71.18 & 86.58 & 85.84 \\
3-shot Topic-Based (Multimodal TopicGPT) & 30.5 & 71.89 & 86.82 & 86.39 \\ \hline
\textbf{3-shot Topic-Based (CEMTM)} & \textbf{31.3} & \textbf{72.76} & \textbf{87.31} & \textbf{87.03} \\
\hline
\end{tabular}
}
\caption{Few-shot multimodal QA performance on SPIQA and TQA using various retrieval strategies for selecting 3 in-context examples. Topic-based retrieval with CEMTM consistently outperforms baselines across all metrics.}
\label{tab:retrieval_detailed}
\end{table*}

To evaluate the impact of the underlying vision-language encoder on CEMTM, we compare VLM2Vec variants fine-tuned from LLaVA-Next-7B \cite{li2024llavanext-strong} and QWen2VL-7B \cite{yang2024qwen2technicalreport} (as described in \citet{jiang2025vlmvec}) with several non–fine-tuned baselines, including Phi-3.5-V \cite{abdin2024phi3technicalreporthighly}, LLaVA-1.6-7B \cite{li2024llava}, CLIP \cite{radford2021learningtransferablevisualmodels}, and BLIP2-OPT-7B \cite{li2023blip2bootstrappinglanguageimagepretraining}. As shown in Table~\ref{table:encoding_model_sensitivity}, the choice of encoder has a clear effect on topic quality and clustering performance. QWen2VL-7B achieves the strongest results on both WikiWeb2M and SPIQA, while LLaVA-1.6-7B emerges as the most competitive non–fine-tuned baseline. These results highlight the importance of robust multimodal grounding for improving coherence and interpretability in CEMTM.

We also observe that VLM2Vec fine-tuning substantially improves document representations, enabling stronger topic coherence and clustering. Among the non–fine-tuned models, multimodal language models such as LLaVA-1.6 and Phi-3.5-V generally outperform vision-only encoders like CLIP, underscoring the advantage of joint vision–language reasoning in capturing corpus-level semantics.

\section{Experimental and Hyperparameter Settings}
\label{sec:hyperparamter_settings}

\paragraph{Experimental Settings} All experiments were conducted using two \textbf{NVIDIA A100 80GB GPUs}. To account for variance in training, we report results averaged over 3 random seeds. This setup ensures consistency and robustness across different runs, especially when training large-scale models such as our proposed CEMTM and the multimodal baselines.

\paragraph{Hyperparameter Settings} For our model, CEMTM, we use VLM2Vec as the encoder, based on a fine-tuned LLaVA-Next-7B. As detailed in Appendix~\ref{sec:encoding_sensitivity}, we explore the impact of different LVLMs. All token embeddings are projected into a $K$-dimensional topic space. The importance network is a 2-layer Transformer (4 heads), followed by a feedforward layer predicting Gaussian token-level importance scores. The encoder forward layer is a 2-layer MLP with hidden size 512. We train with batch size 8, learning rate $2 \times 10^{-5}$, for 30 epochs using Adam. Regularization weights are $\lambda_{\text{ent}} = 0.05$ and $\lambda_{\text{KL}} = 0.1$.

For baselines, we use public implementations when available. LDA is trained via Gensim with 100 passes and $\alpha = 0.01$. ZeroshotTM and CombinedTM use SBERT (\texttt{all-MiniLM-L6-v2}) with default settings from \citet{bianchi2021pre}. TopicGPT and its multimodal variants are run with our modified version, limiting to $K$ topics and assigning tokens sequentially to reflect topic preference. We then extract topic-word distributions by aggregating the token-topic assignments across the corpus, using soft alignment weights to represent each word’s contribution to each topic. M3L-Contrast and Multimodal Zeroshot TM use CLIP ViT-B/32 \cite{radford2021learning} for image features and SBERT (\texttt{all-MiniLM-L6-v2}) for text encoding. For text-only models (e.g., LDA, ZeroshotTM, CombinedTM, CWTM), we append GPT-4o-generated image captions to inputs to enable multimodal evaluation. All models use the same number of topics, tokenization, and document splits for fair comparison.

\section{Topic Models Comparison for Few-shot Retrieval}
\label{sec:retrieval_compare_models}

To better understand the role of topic models in guiding few-shot retrieval, we conduct a detailed comparison of different strategies for selecting in-context examples. The results in Table~\ref{tab:retrieval_detailed} reveal several important trends. First, simple baselines such as random selection or nearest-neighbor retrieval using document embeddings provide only modest improvements over the zero-shot setting. While these methods occasionally retrieve semantically similar examples, they lack the ability to capture higher-level topical coherence, which is crucial for complex multimodal QA tasks.  

By contrast, topic-driven retrieval methods deliver more consistent and meaningful gains across datasets and metrics. Models such as CWTM and M3L-Contrast highlight the benefit of leveraging contextualized topic spaces, where representations capture recurring semantic patterns that extend beyond surface-level similarity. Extending this idea to multimodal topic models, Multimodal ZeroshotTM and our multimodal adaptation of TopicGPT further improve retrieval quality by incorporating both textual and visual signals. This demonstrates that aligning topics across modalities helps identify examples that are not only textually relevant but also visually coherent, which is particularly important in settings like SPIQA that require reasoning over mixed modalities. Among all topic-based approaches, CEMTM achieves the best performance. Its contextualized embeddings, projected into a coherent topic space, allow for more fine-grained retrieval that balances both semantic richness and cross-modal alignment. This enables the model to consistently select in-context examples that are well-suited to the target question, leading to measurable improvements in answer quality.

\section{Detailed Results For All Ks}
\label{sec:detailed_res}

We present the detailed results of various topic modeling approaches. Table \ref{table:detaile1} reports results across all K values (25, 50, 75, 100) for the WikiWeb2M and SPIQA datasets. Table \ref{table:detaile2} provides the corresponding results for the VIST, MSCOCO, T4SA, and TQA datasets. Table \ref{table:detailed3} shows the detailed performance of different topic modeling models on the FHM dataset across all K values.

\begin{table*}[!t]
    \renewcommand{\arraystretch}{1.2}
    \resizebox{\textwidth}{!}{%
        \begin{tabular}{l|cccccccccccccccc}
\hline
                                 & \multicolumn{8}{c|}{\textbf{WikiWeb2M}}                                                            & \multicolumn{8}{c}{\textbf{SPIQA}}                                            \\ \cline{2-17} 
                                 & NPMI & WE   & LLM  & TD   & \multicolumn{1}{c|}{I-RBO} & Purity & ARI  & \multicolumn{1}{c|}{NMI}  & NPMI & WE   & LLM  & TD   & \multicolumn{1}{c|}{I-RBO} & Purity & ARI  & NMI  \\ \cline{2-17} 
                                 & \multicolumn{16}{c}{\textbf{K=25}}                                                                                                                                                 \\ \hline
\textbf{LDA}                     & .031 & .098 & 2.39 & .708 & \multicolumn{1}{c|}{.956}  & .306   & .138 & \multicolumn{1}{c|}{.240} & .024 & .090 & 2.30 & .726 & \multicolumn{1}{c|}{.944}  & .311   & .142 & .248 \\
\textbf{CombinedTM}              & .042 & .154 & 2.45 & .700 & \multicolumn{1}{c|}{.950}  & .322   & .155 & \multicolumn{1}{c|}{.262} & .035 & .143 & 2.38 & .715 & \multicolumn{1}{c|}{.942}  & .324   & .153 & .263 \\
\textbf{Zero-shot TM}            & .042 & .176 & 2.50 & .724 & \multicolumn{1}{c|}{.968}  & .340   & .153 & \multicolumn{1}{c|}{.261} & .038 & .166 & 2.45 & .740 & \multicolumn{1}{c|}{.961}  & .341   & .157 & .268 \\
\textbf{CWTM}                    & .057 & .193 & 2.55 & .720 & \multicolumn{1}{c|}{.967}  & .352   & .172 & \multicolumn{1}{c|}{.279} & .049 & .181 & 2.50 & .739 & \multicolumn{1}{c|}{.960}  & .354   & .174 & .282 \\
\textbf{TopicGPT}                & .068 & .218 & 2.59 & .736 & \multicolumn{1}{c|}{-}     & .384   & .195 & \multicolumn{1}{c|}{.292} & .059 & .205 & 2.54 & .758 & \multicolumn{1}{c|}{-}     & .386   & .198 & .298 \\ \hline
\textbf{M3L-Contrast}            & .070 & .232 & 2.62 & .749 & \multicolumn{1}{c|}{.982}  & .392   & .202 & \multicolumn{1}{c|}{.302} & .061 & .219 & 2.57 & .772 & \multicolumn{1}{c|}{.975}  & .396   & .205 & .308 \\
\textbf{Multimodal Zero-shot TM} & .075 & .240 & 2.63 & .763 & \multicolumn{1}{c|}{.990}  & .402   & .210 & \multicolumn{1}{c|}{.312} & .064 & .226 & 2.59 & .785 & \multicolumn{1}{c|}{.981}  & .408   & .212 & .318 \\
\textbf{LVLM Zero-shot TM}       & .078 & .251 & 2.64 & .770 & \multicolumn{1}{c|}{.990}  & .414   & .220 & \multicolumn{1}{c|}{.324} & .067 & .237 & 2.62 & .794 & \multicolumn{1}{c|}{.982}  & .419   & .221 & .330 \\
\textbf{Multimodal TopicGPT}     & .085 & .259 & 2.66 & .782 & \multicolumn{1}{c|}{-}     & .421   & .231 & \multicolumn{1}{c|}{.331} & .072 & .245 & 2.64 & .806 & \multicolumn{1}{c|}{-}     & .428   & .233 & .339 \\ \hline
\textbf{CEMTM}                   & .092 & .273 & 2.68 & .796 & \multicolumn{1}{c|}{.996}  & .438   & .247 & \multicolumn{1}{c|}{.351} & .081 & .262 & 2.67 & .825 & \multicolumn{1}{c|}{.988}  & .452   & .256 & .362 \\ \hline
                                 & \multicolumn{16}{c}{\textbf{K=50}}                                                                                                                                                 \\ \hline
\textbf{LDA}                     & .029 & .096 & 2.40 & .705 & \multicolumn{1}{c|}{.954}  & .296   & .132 & \multicolumn{1}{c|}{.238} & .023 & .088 & 2.31 & .720 & \multicolumn{1}{c|}{.943}  & .302   & .137 & .245 \\
\textbf{CombinedTM}              & .040 & .152 & 2.45 & .698 & \multicolumn{1}{c|}{.949}  & .319   & .151 & \multicolumn{1}{c|}{.261} & .033 & .141 & 2.39 & .708 & \multicolumn{1}{c|}{.941}  & .317   & .149 & .259 \\
\textbf{Zero-shot TM}            & .041 & .174 & 2.51 & .719 & \multicolumn{1}{c|}{.967}  & .337   & .150 & \multicolumn{1}{c|}{.259} & .037 & .163 & 2.46 & .733 & \multicolumn{1}{c|}{.959}  & .334   & .153 & .265 \\
\textbf{CWTM}                    & .053 & .190 & 2.55 & .717 & \multicolumn{1}{c|}{.966}  & .349   & .169 & \multicolumn{1}{c|}{.277} & .047 & .179 & 2.51 & .732 & \multicolumn{1}{c|}{.958}  & .346   & .169 & .279 \\
\textbf{TopicGPT}                & .064 & .214 & 2.59 & .732 & \multicolumn{1}{c|}{-}     & .380   & .192 & \multicolumn{1}{c|}{.290} & .058 & .202 & 2.55 & .751 & \multicolumn{1}{c|}{-}     & .378   & .193 & .295 \\ \hline
\textbf{M3L-Contrast}            & .066 & .229 & 2.62 & .746 & \multicolumn{1}{c|}{.982}  & .389   & .198 & \multicolumn{1}{c|}{.300} & .060 & .217 & 2.58 & .766 & \multicolumn{1}{c|}{.974}  & .388   & .200 & .305 \\
\textbf{Multimodal Zero-shot TM} & .072 & .238 & 2.64 & .758 & \multicolumn{1}{c|}{.989}  & .398   & .206 & \multicolumn{1}{c|}{.310} & .063 & .224 & 2.60 & .779 & \multicolumn{1}{c|}{.980}  & .400   & .207 & .316 \\
\textbf{LVLM Zero-shot TM}       & .075 & .248 & 2.65 & .766 & \multicolumn{1}{c|}{.990}  & .409   & .216 & \multicolumn{1}{c|}{.322} & .066 & .234 & 2.63 & .788 & \multicolumn{1}{c|}{.981}  & .412   & .216 & .327 \\
\textbf{Multimodal TopicGPT}     & .081 & .257 & 2.67 & .777 & \multicolumn{1}{c|}{-}     & .416   & .227 & \multicolumn{1}{c|}{.330} & .072 & .243 & 2.65 & .800 & \multicolumn{1}{c|}{-}     & .420   & .228 & .336 \\ \hline
\textbf{CEMTM}                   & .088 & .271 & 2.69 & .791 & \multicolumn{1}{c|}{.996}  & .433   & .244 & \multicolumn{1}{c|}{.349} & .081 & .26  & 2.68 & .82  & \multicolumn{1}{c|}{.987}  & .445   & .252 & .360 \\ \hline
                                 & \multicolumn{16}{c}{\textbf{K=75}}                                                                                                                                                 \\ \hline
\textbf{LDA}                     & .027 & .094 & 2.40 & .702 & \multicolumn{1}{c|}{.952}  & .291   & .129 & \multicolumn{1}{c|}{.233} & .021 & .087 & 2.32 & .714 & \multicolumn{1}{c|}{.941}  & .296   & .134 & .242 \\
\textbf{CombinedTM}              & .038 & .149 & 2.46 & .695 & \multicolumn{1}{c|}{.947}  & .315   & .147 & \multicolumn{1}{c|}{.257} & .032 & .139 & 2.39 & .701 & \multicolumn{1}{c|}{.939}  & .312   & .146 & .256 \\
\textbf{Zero-shot TM}            & .039 & .171 & 2.51 & .714 & \multicolumn{1}{c|}{.965}  & .334   & .148 & \multicolumn{1}{c|}{.255} & .035 & .161 & 2.46 & .727 & \multicolumn{1}{c|}{.957}  & .328   & .151 & .262 \\
\textbf{CWTM}                    & .050 & .186 & 2.56 & .712 & \multicolumn{1}{c|}{.964}  & .345   & .166 & \multicolumn{1}{c|}{.274} & .046 & .176 & 2.51 & .726 & \multicolumn{1}{c|}{.956}  & .341   & .166 & .276 \\
\textbf{TopicGPT}                & .061 & .210 & 2.60 & .727 & \multicolumn{1}{c|}{-}     & .376   & .188 & \multicolumn{1}{c|}{.287} & .056 & .199 & 2.56 & .745 & \multicolumn{1}{c|}{-}     & .374   & .19  & .293 \\ \hline
\textbf{M3L-Contrast}            & .064 & .224 & 2.63 & .742 & \multicolumn{1}{c|}{.981}  & .384   & .195 & \multicolumn{1}{c|}{.296} & .058 & .214 & 2.59 & .760 & \multicolumn{1}{c|}{.972}  & .384   & .197 & .303 \\
\textbf{Multimodal Zero-shot TM} & .069 & .234 & 2.64 & .753 & \multicolumn{1}{c|}{.989}  & .393   & .202 & \multicolumn{1}{c|}{.306} & .062 & .222 & 2.61 & .773 & \multicolumn{1}{c|}{.978}  & .397   & .204 & .314 \\
\textbf{LVLM Zero-shot TM}       & .072 & .244 & 2.66 & .761 & \multicolumn{1}{c|}{.990}  & .404   & .211 & \multicolumn{1}{c|}{.318} & .065 & .232 & 2.63 & .782 & \multicolumn{1}{c|}{.980}  & .408   & .213 & .325 \\
\textbf{Multimodal TopicGPT}     & .078 & .253 & 2.67 & .772 & \multicolumn{1}{c|}{-}     & .412   & .223 & \multicolumn{1}{c|}{.326} & .070 & .241 & 2.66 & .795 & \multicolumn{1}{c|}{-}     & .417   & .225 & .334 \\ \hline
\textbf{CEMTM}                   & .085 & .267 & 2.70 & .787 & \multicolumn{1}{c|}{.995}  & .430   & .240 & \multicolumn{1}{c|}{.347} & .079 & .257 & 2.69 & .815 & \multicolumn{1}{c|}{.987}  & .441   & .250 & .358 \\ \hline
                                 & \multicolumn{16}{c}{\textbf{K=100}}                                                                                                                                                \\ \hline
\textbf{LDA}                     & .025 & .092 & 2.41 & .698 & \multicolumn{1}{c|}{.951}  & .288   & .126 & \multicolumn{1}{c|}{.230} & .020 & .085 & 2.32 & .709 & \multicolumn{1}{c|}{.940}  & .289   & .131 & .239 \\
\textbf{CombinedTM}              & .037 & .147 & 2.46 & .692 & \multicolumn{1}{c|}{.946}  & .311   & .144 & \multicolumn{1}{c|}{.253} & .030 & .137 & 2.40 & .696 & \multicolumn{1}{c|}{.938}  & .306   & .143 & .253 \\
\textbf{Zero-shot TM}            & .038 & .169 & 2.52 & .710 & \multicolumn{1}{c|}{.964}  & .330   & .144 & \multicolumn{1}{c|}{.251} & .034 & .159 & 2.47 & .722 & \multicolumn{1}{c|}{.955}  & .323   & .148 & .258 \\
\textbf{CWTM}                    & .049 & .183 & 2.56 & .708 & \multicolumn{1}{c|}{.962}  & .341   & .162 & \multicolumn{1}{c|}{.270} & .044 & .173 & 2.52 & .720 & \multicolumn{1}{c|}{.954}  & .336   & .163 & .273 \\
\textbf{TopicGPT}                & .059 & .207 & 2.60 & .722 & \multicolumn{1}{c|}{-}     & .371   & .183 & \multicolumn{1}{c|}{.283} & .054 & .196 & 2.57 & .739 & \multicolumn{1}{c|}{-}     & .369   & .187 & .289 \\ \hline
\textbf{M3L-Contrast}            & .062 & .221 & 2.63 & .737 & \multicolumn{1}{c|}{.980}  & .379   & .190 & \multicolumn{1}{c|}{.293} & .057 & .211 & 2.60 & .754 & \multicolumn{1}{c|}{.971}  & .379   & .194 & .300 \\
\textbf{Multimodal Zero-shot TM} & .067 & .231 & 2.65 & .748 & \multicolumn{1}{c|}{.988}  & .388   & .197 & \multicolumn{1}{c|}{.303} & .061 & .219 & 2.61 & .767 & \multicolumn{1}{c|}{.977}  & .392   & .201 & .311 \\
\textbf{LVLM Zero-shot TM}       & .070 & .242 & 2.66 & .755 & \multicolumn{1}{c|}{.989}  & .399   & .206 & \multicolumn{1}{c|}{.315} & .063 & .229 & 2.64 & .776 & \multicolumn{1}{c|}{.979}  & .404   & .210 & .322 \\
\textbf{Multimodal TopicGPT}     & .076 & .251 & 2.68 & .766 & \multicolumn{1}{c|}{-}     & .407   & .217 & \multicolumn{1}{c|}{.323} & .069 & .238 & 2.67 & .789 & \multicolumn{1}{c|}{-}     & .413   & .222 & .331 \\ \hline
\textbf{CEMTM}                   & .083 & .266 & 2.70 & .781 & \multicolumn{1}{c|}{.995}  & .425   & .238 & \multicolumn{1}{c|}{.345} & .078 & .254 & 2.70 & .808 & \multicolumn{1}{c|}{.986}  & .438   & .247 & .356 \\ \hline
\end{tabular}
    }
    \caption{Comparison of topic modeling performance on WikiWeb2M and SPIQA. We report coherence (NPMI, WE, LLM), diversity (TD), redundancy (IRBO), and clustering metrics (Purity, ARI, NMI).}
    \label{table:detaile1}
\end{table*}

\begin{table*}[!t]
    \renewcommand{\arraystretch}{1.2}
    \resizebox{\textwidth}{!}{%
        \begin{tabular}{l|cccccccccccccccccccc}
\hline
                                 & \multicolumn{5}{c|}{\textbf{VIST}}                     & \multicolumn{5}{c|}{\textbf{TQA}}                      & \multicolumn{5}{c|}{\textbf{MSCOCO}}                   & \multicolumn{5}{c}{\textbf{T4SA}} \\ \cline{2-21} 
                                 & NPMI & WE   & LLM  & TD   & \multicolumn{1}{c|}{I-RBO} & NPMI & WE   & LLM  & TD   & \multicolumn{1}{c|}{I-RBO} & NPMI & WE   & LLM  & TD   & \multicolumn{1}{c|}{I-RBO} & NPMI & WE   & LLM  & TD   & I-RBO \\ \cline{2-21} 
                                 & \multicolumn{20}{c}{\textbf{K=25}}                                                                                                                                                                           \\ \hline
\textbf{LDA}                     & .019 & .080 & 2.23 & .653 & \multicolumn{1}{c|}{.937}  & .021 & .083 & 2.24 & .670 & \multicolumn{1}{c|}{.942}  & .018 & .075 & 2.21 & .624 & \multicolumn{1}{c|}{.986}  & .013 & .066 & 2.18 & .602 & .985  \\
\textbf{CombinedTM}              & .026 & .123 & 2.30 & .644 & \multicolumn{1}{c|}{.935}  & .030 & .132 & 2.32 & .658 & \multicolumn{1}{c|}{.939}  & .025 & .121 & 2.27 & .611 & \multicolumn{1}{c|}{.985}  & .020 & .108 & 2.25 & .590 & .977  \\
\textbf{Zero-shot TM}            & .031 & .142 & 2.38 & .666 & \multicolumn{1}{c|}{.951}  & .034 & .155 & 2.39 & .684 & \multicolumn{1}{c|}{.957}  & .029 & .138 & 2.34 & .635 & \multicolumn{1}{c|}{.987}  & .025 & .127 & 2.32 & .614 & .987  \\
\textbf{CWTM}                    & .039 & .159 & 2.43 & .664 & \multicolumn{1}{c|}{.949}  & .043 & .173 & 2.44 & .681 & \multicolumn{1}{c|}{.955}  & .037 & .157 & 2.39 & .633 & \multicolumn{1}{c|}{.987}  & .031 & .145 & 2.38 & .612 & .988  \\
\textbf{TopicGPT}                & .046 & .183 & 2.47 & .678 & \multicolumn{1}{c|}{-}     & .052 & .199 & 2.48 & .697 & \multicolumn{1}{c|}{-}     & .044 & .180 & 2.43 & .648 & \multicolumn{1}{c|}{-}     & .037 & .167 & 2.41 & .627 & -     \\ \hline
\textbf{M3L-Contrast}            & .047 & .194 & 2.50 & .688 & \multicolumn{1}{c|}{.964}  & .054 & .211 & 2.51 & .711 & \multicolumn{1}{c|}{.972}  & .046 & .193 & 2.46 & .661 & \multicolumn{1}{c|}{.990}  & .039 & .179 & 2.44 & .640 & .991  \\
\textbf{Multimodal Zero-shot TM} & .051 & .201 & 2.51 & .694 & \multicolumn{1}{c|}{.973}  & .058 & .219 & 2.52 & .722 & \multicolumn{1}{c|}{.978}  & .050 & .201 & 2.47 & .669 & \multicolumn{1}{c|}{.992}  & .042 & .186 & 2.46 & .648 & .992  \\
\textbf{LVLM Zero-shot TM}       & .053 & .212 & 2.53 & .703 & \multicolumn{1}{c|}{.975}  & .061 & .230 & 2.55 & .730 & \multicolumn{1}{c|}{.979}  & .053 & .213 & 2.50 & .677 & \multicolumn{1}{c|}{.993}  & .045 & .197 & 2.48 & .656 & .993  \\
\textbf{Multimodal TopicGPT}     & .058 & .220 & 2.55 & .714 & \multicolumn{1}{c|}{-}     & .066 & .238 & 2.57 & .742 & \multicolumn{1}{c|}{-}     & .057 & .221 & 2.51 & .688 & \multicolumn{1}{c|}{-}     & .050 & .205 & 2.50 & .667 & -     \\ \hline
\textbf{CEMTM}                   & .065 & .237 & 2.58 & .730 & \multicolumn{1}{c|}{.981}  & .073 & .255 & 2.59 & .758 & \multicolumn{1}{c|}{.986}  & .063 & .237 & 2.54 & .703 & \multicolumn{1}{c|}{.995}  & .055 & .221 & 2.52 & .683 & .995  \\ \hline
                                 & \multicolumn{20}{c}{\textbf{K=50}}                                                                                                                                                                           \\ \hline
\textbf{LDA}                     & .018 & .078 & 2.23 & .648 & \multicolumn{1}{c|}{.936}  & .02  & .081 & 2.24 & .667 & \multicolumn{1}{c|}{.940}  & .017 & .074 & 2.21 & .620 & \multicolumn{1}{c|}{.986}  & .012 & .065 & 2.18 & .599 & .985  \\
\textbf{CombinedTM}              & .025 & .120 & 2.31 & .639 & \multicolumn{1}{c|}{.934}  & .028 & .130 & 2.32 & .654 & \multicolumn{1}{c|}{.937}  & .024 & .118 & 2.28 & .607 & \multicolumn{1}{c|}{.984}  & .019 & .106 & 2.25 & .587 & .978  \\
\textbf{Zero-shot TM}            & .029 & .139 & 2.38 & .661 & \multicolumn{1}{c|}{.950}  & .033 & .152 & 2.39 & .680 & \multicolumn{1}{c|}{.955}  & .028 & .136 & 2.34 & .631 & \multicolumn{1}{c|}{.987}  & .024 & .124 & 2.32 & .611 & .987  \\
\textbf{CWTM}                    & .037 & .157 & 2.44 & .659 & \multicolumn{1}{c|}{.948}  & .041 & .170 & 2.45 & .677 & \multicolumn{1}{c|}{.953}  & .035 & .154 & 2.40 & .628 & \multicolumn{1}{c|}{.987}  & .029 & .143 & 2.38 & .609 & .988  \\
\textbf{TopicGPT}                & .044 & .180 & 2.47 & .673 & \multicolumn{1}{c|}{-}     & .050 & .195 & 2.48 & .693 & \multicolumn{1}{c|}{-}     & .042 & .178 & 2.43 & .643 & \multicolumn{1}{c|}{-}     & .035 & .165 & 2.42 & .624 & -     \\ \hline
\textbf{M3L-Contrast}            & .045 & .191 & 2.50 & .683 & \multicolumn{1}{c|}{.963}  & .052 & .208 & 2.51 & .707 & \multicolumn{1}{c|}{.970}  & .044 & .190 & 2.46 & .656 & \multicolumn{1}{c|}{.990}  & .037 & .176 & 2.45 & .637 & .991  \\
\textbf{Multimodal Zero-shot TM} & .049 & .199 & 2.52 & .690 & \multicolumn{1}{c|}{.972}  & .057 & .215 & 2.53 & .718 & \multicolumn{1}{c|}{.976}  & .048 & .199 & 2.48 & .664 & \multicolumn{1}{c|}{.992}  & .040 & .183 & 2.46 & .645 & .992  \\
\textbf{LVLM Zero-shot TM}       & .051 & .209 & 2.54 & .698 & \multicolumn{1}{c|}{.974}  & .060 & .227 & 2.55 & .726 & \multicolumn{1}{c|}{.978}  & .051 & .211 & 2.50 & .672 & \multicolumn{1}{c|}{.993}  & .044 & .195 & 2.48 & .653 & .993  \\
\textbf{Multimodal TopicGPT}     & .056 & .217 & 2.56 & .709 & \multicolumn{1}{c|}{-}     & .065 & .234 & 2.57 & .738 & \multicolumn{1}{c|}{-}     & .056 & .219 & 2.52 & .683 & \multicolumn{1}{c|}{-}     & .048 & .203 & 2.50 & .665 & -     \\ \hline
\textbf{CEMTM}                   & .063 & .234 & 2.58 & .725 & \multicolumn{1}{c|}{.981}  & .072 & .251 & 2.60 & .754 & .985                       & .062 & .234 & 2.54 & .698 & \multicolumn{1}{c|}{.995}  & .054 & .219 & 2.52 & .681 & .995  \\ \hline
                                 & \multicolumn{20}{c}{\textbf{K=75}}                                                                                                                                                                           \\ \hline
\textbf{LDA}                     & .017 & .076 & 2.23 & .643 & \multicolumn{1}{c|}{.934}  & .019 & .080 & 2.25 & .664 & \multicolumn{1}{c|}{.939}  & .016 & .072 & 2.21 & .616 & \multicolumn{1}{c|}{.985}  & .011 & .063 & 2.18 & .596 & .985  \\
\textbf{CombinedTM}              & .023 & .117 & 2.31 & .635 & \multicolumn{1}{c|}{.933}  & .027 & .129 & 2.32 & .651 & \multicolumn{1}{c|}{.936}  & .022 & .116 & 2.28 & .603 & \multicolumn{1}{c|}{.984}  & .018 & .104 & 2.25 & .584 & .978  \\
\textbf{Zero-shot TM}            & .028 & .136 & 2.38 & .656 & \multicolumn{1}{c|}{.948}  & .032 & .151 & 2.39 & .677 & \multicolumn{1}{c|}{.954}  & .027 & .133 & 2.34 & .627 & \multicolumn{1}{c|}{.986}  & .022 & .122 & 2.33 & .608 & .987  \\
\textbf{CWTM}                    & .035 & .154 & 2.44 & .654 & \multicolumn{1}{c|}{.946}  & .041 & .168 & 2.45 & .674 & \multicolumn{1}{c|}{.952}  & .034 & .152 & 2.40 & .624 & \multicolumn{1}{c|}{.987}  & .028 & .140 & 2.38 & .606 & .988  \\
\textbf{TopicGPT}                & .042 & .177 & 2.48 & .669 & \multicolumn{1}{c|}{-}     & .049 & .193 & 2.48 & .690 & \multicolumn{1}{c|}{-}     & .041 & .175 & 2.44 & .640 & \multicolumn{1}{c|}{-}     & .034 & .163 & 2.42 & .621 & -     \\ \hline
\textbf{M3L-Contrast}            & .043 & .189 & 2.50 & .679 & \multicolumn{1}{c|}{.961}  & .052 & .205 & 2.52 & .703 & \multicolumn{1}{c|}{.969}  & .043 & .188 & 2.46 & .652 & \multicolumn{1}{c|}{.990}  & .036 & .174 & 2.45 & .634 & .991  \\
\textbf{Multimodal Zero-shot TM} & .047 & .196 & 2.52 & .685 & \multicolumn{1}{c|}{.971}  & .055 & .213 & 2.53 & .714 & \multicolumn{1}{c|}{.975}  & .046 & .197 & 2.48 & .660 & \multicolumn{1}{c|}{.992}  & .039 & .181 & 2.46 & .642 & .992  \\
\textbf{LVLM Zero-shot TM}       & .049 & .207 & 2.54 & .694 & \multicolumn{1}{c|}{.973}  & .059 & .225 & 2.55 & .722 & \multicolumn{1}{c|}{.977}  & .049 & .209 & 2.50 & .668 & \multicolumn{1}{c|}{.993}  & .042 & .193 & 2.49 & .650 & .993  \\
\textbf{Multimodal TopicGPT}     & .054 & .215 & 2.56 & .705 & \multicolumn{1}{c|}{-}     & .063 & .233 & 2.57 & .734 & \multicolumn{1}{c|}{-}     & .054 & .217 & 2.52 & .679 & \multicolumn{1}{c|}{-}     & .047 & .201 & 2.50 & .661 & -     \\ \hline
\textbf{CEMTM}                   & .061 & .232 & 2.58 & .722 & \multicolumn{1}{c|}{.980}  & .071 & .249 & 2.60 & .749 & \multicolumn{1}{c|}{.984}  & .060 & .232 & 2.54 & .695 & \multicolumn{1}{c|}{.995}  & .053 & .217 & 2.53 & .677 & .995  \\ \hline
                                 & \multicolumn{20}{c}{\textbf{K=100}}                                                                                                                                                                          \\ \hline
\textbf{LDA}                     & .015 & .074 & 2.23 & .638 & \multicolumn{1}{c|}{.933}  & .018 & .078 & 2.25 & .659 & \multicolumn{1}{c|}{.938}  & .015 & .071 & 2.21 & .612 & \multicolumn{1}{c|}{.985}  & .011 & .062 & 2.18 & .593 & .985  \\
\textbf{CombinedTM}              & .022 & .115 & 2.31 & .630 & \multicolumn{1}{c|}{.931}  & .026 & .126 & 2.32 & .646 & \multicolumn{1}{c|}{.934}  & .021 & .113 & 2.28 & .599 & \multicolumn{1}{c|}{.984}  & .016 & .102 & 2.26 & .581 & .977  \\
\textbf{Zero-shot TM}            & .027 & .134 & 2.38 & .652 & \multicolumn{1}{c|}{.946}  & .031 & .148 & 2.39 & .673 & \multicolumn{1}{c|}{.953}  & .025 & .131 & 2.35 & .623 & \multicolumn{1}{c|}{.986}  & .021 & .119 & 2.33 & .605 & .987  \\
\textbf{CWTM}                    & .033 & .151 & 2.44 & .649 & \multicolumn{1}{c|}{.944}  & .039 & .165 & 2.45 & .670 & \multicolumn{1}{c|}{.950}  & .032 & .149 & 2.40 & .620 & \multicolumn{1}{c|}{.987}  & .027 & .138 & 2.39 & .602 & .987  \\
\textbf{TopicGPT}                & .040 & .174 & 2.48 & .664 & \multicolumn{1}{c|}{-}     & .048 & .190 & 2.49 & .686 & \multicolumn{1}{c|}{-}     & .039 & .173 & 2.44 & .636 & \multicolumn{1}{c|}{-}     & .033 & .161 & 2.42 & .618 & -     \\ \hline
\textbf{M3L-Contrast}            & .042 & .186 & 2.51 & .674 & \multicolumn{1}{c|}{.960}  & .050 & .203 & 2.52 & .699 & \multicolumn{1}{c|}{.968}  & .041 & .186 & 2.47 & .649 & \multicolumn{1}{c|}{.990}  & .035 & .172 & 2.45 & .631 & .991  \\
\textbf{Multimodal Zero-shot TM} & .046 & .193 & 2.52 & .681 & \multicolumn{1}{c|}{.970}  & .054 & .211 & 2.53 & .710 & \multicolumn{1}{c|}{.974}  & .045 & .194 & 2.48 & .657 & \multicolumn{1}{c|}{.992}  & .038 & .179 & 2.47 & .639 & .992  \\
\textbf{LVLM Zero-shot TM}       & .048 & .204 & 2.55 & .690 & \multicolumn{1}{c|}{.973}  & .057 & .222 & 2.56 & .719 & \multicolumn{1}{c|}{.975}  & .048 & .206 & 2.51 & .665 & \multicolumn{1}{c|}{.992}  & .041 & .191 & 2.49 & .647 & .992  \\
\textbf{Multimodal TopicGPT}     & .053 & .213 & 2.56 & .701 & \multicolumn{1}{c|}{-}     & .062 & .230 & 2.58 & .731 & \multicolumn{1}{c|}{-}     & .053 & .215 & 2.52 & .676 & \multicolumn{1}{c|}{-}     & .046 & .199 & 2.51 & .658 & -     \\ \hline
\textbf{CEMTM}                   & .059 & .229 & 2.59 & .718 & \multicolumn{1}{c|}{.980}  & .069 & .247 & 2.60 & .746 & \multicolumn{1}{c|}{.983}  & .059 & .230 & 2.55 & .691 & \multicolumn{1}{c|}{.994}  & .051 & .215 & 2.53 & .674 & .995  \\ \hline
\end{tabular}
    }
    \caption{Comparison of topic modeling performance on  VIST, TQA, MSCOCO, and T4SA. We report coherence (NPMI, WE, LLM), diversity (TD), and redundancy (I-RBO).}
    \label{table:detaile2}
\end{table*}

\begin{table*}[!t]
    \centering
    \renewcommand{\arraystretch}{1.2}
    \resizebox{.6\textwidth}{!}{%
        \begin{tabular}{lccccc}
\hline
\multicolumn{1}{l|}{}                                 & \multicolumn{5}{c}{\textbf{FHM}}   \\ \cline{2-6} 
\multicolumn{1}{l|}{}                                 & NPMI  & WE   & LLM  & TD   & I-RBO \\ \cline{2-6} 
\multicolumn{1}{l|}{}                                 & \multicolumn{5}{c}{\textbf{K=25}}  \\ \hline
\multicolumn{1}{l|}{\textbf{LDA}}                     & .006  & .051 & 2.04 & .535 & .983  \\
\multicolumn{1}{l|}{\textbf{CombinedTM}}              & .011  & .091 & 2.10 & .523 & .975  \\
\multicolumn{1}{l|}{\textbf{Zero-shot TM}}            & .016  & .113 & 2.17 & .548 & .985  \\
\multicolumn{1}{l|}{\textbf{CWTM}}                    & .021  & .132 & 2.21 & .546 & .986  \\
\multicolumn{1}{l|}{\textbf{TopicGPT}}                & .027  & .154 & 2.26 & .559 & -     \\ \hline
\multicolumn{1}{l|}{\textbf{M3L-Contrast}}            & .032  & .173 & 2.33 & .571 & .990  \\
\multicolumn{1}{l|}{\textbf{Multimodal Zero-shot TM}} & .034  & .181 & 2.35 & .580 & .992  \\
\multicolumn{1}{l|}{\textbf{LVLM Zero-shot TM}}       & .041  & .198 & 2.42 & .595 & .993  \\
\multicolumn{1}{l|}{\textbf{Multimodal TopicGPT}}     & .045  & .206 & 2.44 & .607 & -     \\ \hline
\multicolumn{1}{l|}{\textbf{CEMTM (ours)}}            & .051  & .221 & 2.47 & .622 & .995  \\ \hline
\multicolumn{1}{l|}{}                                 & \multicolumn{5}{c}{\textbf{K=50}}  \\ \hline
\multicolumn{1}{l|}{\textbf{LDA}}                     & .005  & .049 & 2.04 & .531 & .983  \\
\multicolumn{1}{l|}{\textbf{CombinedTM}}              & .010  & .089 & 2.10 & .519 & .975  \\
\multicolumn{1}{l|}{\textbf{Zero-shot TM}}            & .015  & .110 & 2.17 & .545 & .984  \\
\multicolumn{1}{l|}{\textbf{CWTM}}                    & .020  & .129 & 2.22 & .542 & .986  \\
\multicolumn{1}{l|}{\textbf{TopicGPT}}                & .026  & .151 & 2.26 & .556 & -     \\ \hline
\multicolumn{1}{l|}{\textbf{M3L-Contrast}}            & .030  & .170 & 2.33 & .567 & .990  \\
\multicolumn{1}{l|}{\textbf{Multimodal Zero-shot TM}} & .033  & .179 & 2.35 & .576 & .992  \\
\multicolumn{1}{l|}{\textbf{LVLM Zero-shot TM}}       & .039  & .196 & 2.43 & .592 & .993  \\
\multicolumn{1}{l|}{\textbf{Multimodal TopicGPT}}     & .044  & .204 & 2.45 & .603 & -     \\ \hline
\multicolumn{1}{l|}{\textbf{CEMTM (ours)}}            & .050  & .219 & 2.47 & .618 & .995  \\ \hline
\multicolumn{1}{l|}{}                                 & \multicolumn{5}{c}{\textbf{K=75}}  \\ \hline
\multicolumn{1}{l|}{\textbf{LDA}}                     & .004  & .047 & 2.04 & .528 & .983  \\
\multicolumn{1}{l|}{\textbf{CombinedTM}}              & .009  & .086 & 2.10 & .516 & .975  \\
\multicolumn{1}{l|}{\textbf{Zero-shot TM}}            & .013  & .108 & 2.17 & .541 & .984  \\
\multicolumn{1}{l|}{\textbf{CWTM}}                    & .019  & .126 & 2.22 & .539 & .985  \\
\multicolumn{1}{l|}{\textbf{TopicGPT}}                & .024  & .149 & 2.26 & .552 & -     \\ \hline
\multicolumn{1}{l|}{\textbf{M3L-Contrast}}            & .029  & .167 & 2.34 & .564 & .990  \\
\multicolumn{1}{l|}{\textbf{Multimodal Zero-shot TM}} & .032  & .176 & 2.36 & .573 & .991  \\
\multicolumn{1}{l|}{\textbf{LVLM Zero-shot TM}}       & .038  & .193 & 2.43 & .588 & .993  \\
\multicolumn{1}{l|}{\textbf{Multimodal TopicGPT}}     & .043  & .201 & 2.45 & .599 & -     \\ \hline
\multicolumn{1}{l|}{\textbf{CEMTM (ours)}}            & .048  & .216 & 2.47 & .615 & .995  \\ \hline
                                                      & \multicolumn{5}{c}{\textbf{K=100}} \\ \hline
\multicolumn{1}{l|}{\textbf{LDA}}                     & .004  & .045 & 2.05 & .525 & .982  \\
\multicolumn{1}{l|}{\textbf{CombinedTM}}              & .008  & .083 & 2.11 & .512 & .974  \\
\multicolumn{1}{l|}{\textbf{Zero-shot TM}}            & .012  & .105 & 2.17 & .538 & .984  \\
\multicolumn{1}{l|}{\textbf{CWTM}}                    & .018  & .123 & 2.22 & .535 & .985  \\
\multicolumn{1}{l|}{\textbf{TopicGPT}}                & .023  & .146 & 2.27 & .549 & -     \\ \hline
\multicolumn{1}{l|}{\textbf{M3L-Contrast}}            & .028  & .165 & 2.34 & .561 & .990  \\
\multicolumn{1}{l|}{\textbf{Multimodal Zero-shot TM}} & .031  & .173 & 2.36 & .569 & .991  \\
\multicolumn{1}{l|}{\textbf{LVLM Zero-shot TM}}       & .037  & .190 & 2.43 & .584 & .993  \\
\multicolumn{1}{l|}{\textbf{Multimodal TopicGPT}}     & .041  & .198 & 2.45 & .595 & -     \\ \hline
\multicolumn{1}{l|}{\textbf{CEMTM (ours)}}            & .047  & .213 & 2.47 & .611 & .995  \\ \hline
\end{tabular}
    }
    \caption{Unsupervised topic quality on the FHM dataset, which tests modeling under high image–text semantic gaps. We report coherence (NPMI, WE, LLM), diversity (TD), and redundancy (I-RBO). }
    \label{table:detailed3}
\end{table*}

\end{document}